%% file: main.tex
\documentclass[sigconf]{acmart}

\AtBeginDocument{%
  \providecommand\BibTeX{{%
    \normalfont B\kern-0.5em{\scshape i\kern-0.25em b}\kern-0.8em\TeX}}}

\usepackage{bbm}
\usepackage{dsfont}
\usepackage{bm}
\usepackage{booktabs}
\usepackage{subcaption}
\usepackage{amsfonts,amsthm}  
\usepackage{nicefrac}
\usepackage{xcolor}
\usepackage{float}
\usepackage{wrapfig}

\copyrightyear{2022}
\acmYear{2022}
\setcopyright{acmcopyright}\acmConference[WWW '22]{Proceedings of the ACM Web Conference 2022}{April 25--29, 2022}{Virtual Event, Lyon, France}
\acmBooktitle{Proceedings of the ACM Web Conference 2022 (WWW '22), April 25--29, 2022, Virtual Event, Lyon, France}
\acmPrice{15.00}
\acmDOI{10.1145/3485447.3512200}
\acmISBN{978-1-4503-9096-5/22/04}
\newcommand\mypar[1]{\noindent \textit{#1}}

\begin{document}

\title{Augmentations in Graph Contrastive Learning: Current Methodological Flaws \& Towards Better Practices}
\author{Puja Trivedi}
\affiliation{%
  \institution{University of Michigan}
  \country{}}
\email{pujat@umich.edu}
\author{Ekdeep Singh Lubana}
\affiliation{%
  \institution{University of Michigan}
  \country{}
  }
\email{eslubana@umich.edu}
\author{Yujun Yan}
\affiliation{%
  \institution{University of Michigan}
  \country{}
  }
\email{yujunyan@umich.edu}
\author{Yaoqing Yang}
\affiliation{%
  \institution{University of California, Berkeley}
  \country{}
  }
\email{yqyang@berkeley.edu}
\author{Danai Koutra}
\affiliation{%
  \institution{University of Michigan}
  \country{}
  }
\email{dkoutra@umich.edu}




\begin{abstract}
\input{abstract}
\end{abstract}




\maketitle
\section{Introduction}
\label{sec:introduction}
\input{PAGES/010_introduction_v3}
\section{Preliminaries \& Related Work}
\label{sec:preliminaries}
\input{PAGES/020_preliminaries_v3}
\section{Revisiting Augmentations \& Evaluation in GCL}
\label{sec:understanding_graph_cl}
\input{PAGES/030_revisiting_aug}
\section{Benefits \& Design of Task-Aware Augmentations}
\label{sec:effective_augmentations}
\input{PAGES/040_effective_augs_v2}
\section{Conclusion}
\label{sec:conclusion}
\input{PAGES/050_conclusion}
\begin{acks}
We thank Jay Thiagarajan and Mark Heimann for helpful discussions on the project. 
This work is partially supported by the National Science Foundation under CAREER Grant No.~IIS 1845491, Army Young Investigator Award No.~W9-11NF1810397, and Adobe, Amazon, Facebook, and Google faculty awards. 
Any opinions, findings, and conclusions or recommendations expressed here are those of the author(s) and do not reflect the views of funding parties.
\end{acks}

\balance
\bibliographystyle{ACM-Reference-Format}
\bibliography{aaai22.bib}
\clearpage
\appendix
\section{Experimental Details of Section 3}
\input{APPENDIX/ObsAll}

\section{Document Classification}\label{appendex:docu}
\input{APPENDIX/NLP}
\section{Super-pixel Classification}\label{appendex:superpix}
\input{APPENDIX/Superpixel}

\section{Additional Related Work}\label{appendix:related}
\input{APPENDIX/related_work}
\end{document}

%% file: abstract.tex
Unsupervised graph representation learning is critical to a wide range of applications where labels may be scarce or expensive to procure.
Contrastive learning (CL) is an increasingly popular paradigm for such settings and the state-of-the-art in unsupervised \textit{visual} representation learning. 
Recent work attributes the success of visual CL to use of task-relevant augmentations and large, diverse datasets. 
Interestingly, \textit{graph} CL frameworks report strong performance despite using orders of magnitude smaller datasets and employing domain-agnostic graph augmentations (DAGAs).
Motivated by this discrepancy, we probe the quality of representations learnt by popular graph CL frameworks using DAGAs.
We find that DAGAs can destroy task-relevant information and harm the model's ability to learn discriminative representations. On small benchmark datasets, we show the inductive bias of graph neural networks can significantly compensate for this weak discriminability.
Based on our findings, we propose several sanity checks that enable practitioners to quickly assess the quality of their model's learned representations.
We further propose a broad strategy for designing task-aware augmentations that are amenable to graph CL and demonstrate its efficacy on two large-scale, complex graph applications. 
For example, in graph-based document classification, we show task-relevant augmentations improve accuracy up to 20$\%$. 

%% file: PAGES/010_introduction_v3.tex
Graph neural networks (GNNs) have been successfully used to learn representations for various supervised or semi-supervised graph-based tasks, including graph-based similarity search for web documents~\cite{Tao22-webclassification}, fake news detection through propagation pattern classification~\cite{dou2021user,BianXXZHRH20,monti2019fake,HanKL21}, activity analysis in web and social networks (e.g., discussion threads on Reddit, code repository networks on Github)~\cite{karateclub}, and scientific graph classification~\cite{YanZDSSK19,Wale06_NCI1,HeimannSK19}.  
However, in many practical scenarios, labels are scarce or difficult to obtain.
For example, web pages are seldom assigned with labels which summarize their contents, labeling fake news can be time-consuming, and labeling drugs according to their toxicity requires expensive wet lab experiments or analysis~\cite{Duvenaud15_MolFingerprints,Hu20_PretrainingGNNs, Hwang20_SurveyMolecules,Zitnik_Biology}. 
\textit{Contrastive learning} (CL) is an increasingly popular unsupervised graph representation learning paradigm for such label scarce settings~\cite{You20_GraphContrastiveLearning,You21_GCLA,Hassani20_MVGRL,Sun20_InfoGraph,You21_GCLA,Fang21_KAMolecularCL} and is currently the state-of-the-art in unsupervised visual representation learning~\cite{Chen20_SimCLR,He20_MoCo,DBLP:conf/nips/Caron20_Swav,Caron21_DINO}. 

Broadly, CL frameworks learn representations by maximizing similarity between augmentations of a sample (positive views) while simultaneously minimizing similarity to other samples in the batch (negative views).  
Recent theoretical and empirical works attribute the impressive success of visual CL (VCL) to two key principles: (i) leveraging \textit{strong, task-relevant data augmentation}~\cite{Tian20_WhatMakesForGoodViews, Kugelgen_ProvablySeparates, Zimmermann_Inverts, Purushwalkam20_DemystifyingCL, flearning} and (ii) training on \textit{large, diverse datasets} \cite{Chen20_SimCLR, He20_MoCo, MI_bounds, negsamples_1, negsamples_2}. 
By using appropriate data augmentations, VCL frameworks learn high quality representations that are \textit{invariant} to properties \textit{irrelevant} to downstream task performance; thereby preserving task-relevant properties and preventing the model from learning brittle shortcuts~\cite{Tian20_WhatMakesForGoodViews, Purushwalkam20_DemystifyingCL, Chen20_SimCLR,Robinson21_CLShortcuts}. 
Large, diverse datasets are necessary as VCL frameworks routinely use 1K--8K samples in a batch to ensure that enough negative views are available to train stably~\cite{Chen20_SimCLR, He20_MoCo, DBLP:conf/nips/Caron20_Swav, MoCov3}. 
Representations learnt using VCL and self-supervised learning in general have been found to be more robust~\cite{Hendrycks19_SSLOod}, transferable~\cite{Islam21_CLTranserability} and semantically aligned~\cite{Selvaraju20_Casting} than their supervised counterparts.

\begin{figure}[t]
    \centering
    \includegraphics[width=.87\columnwidth]{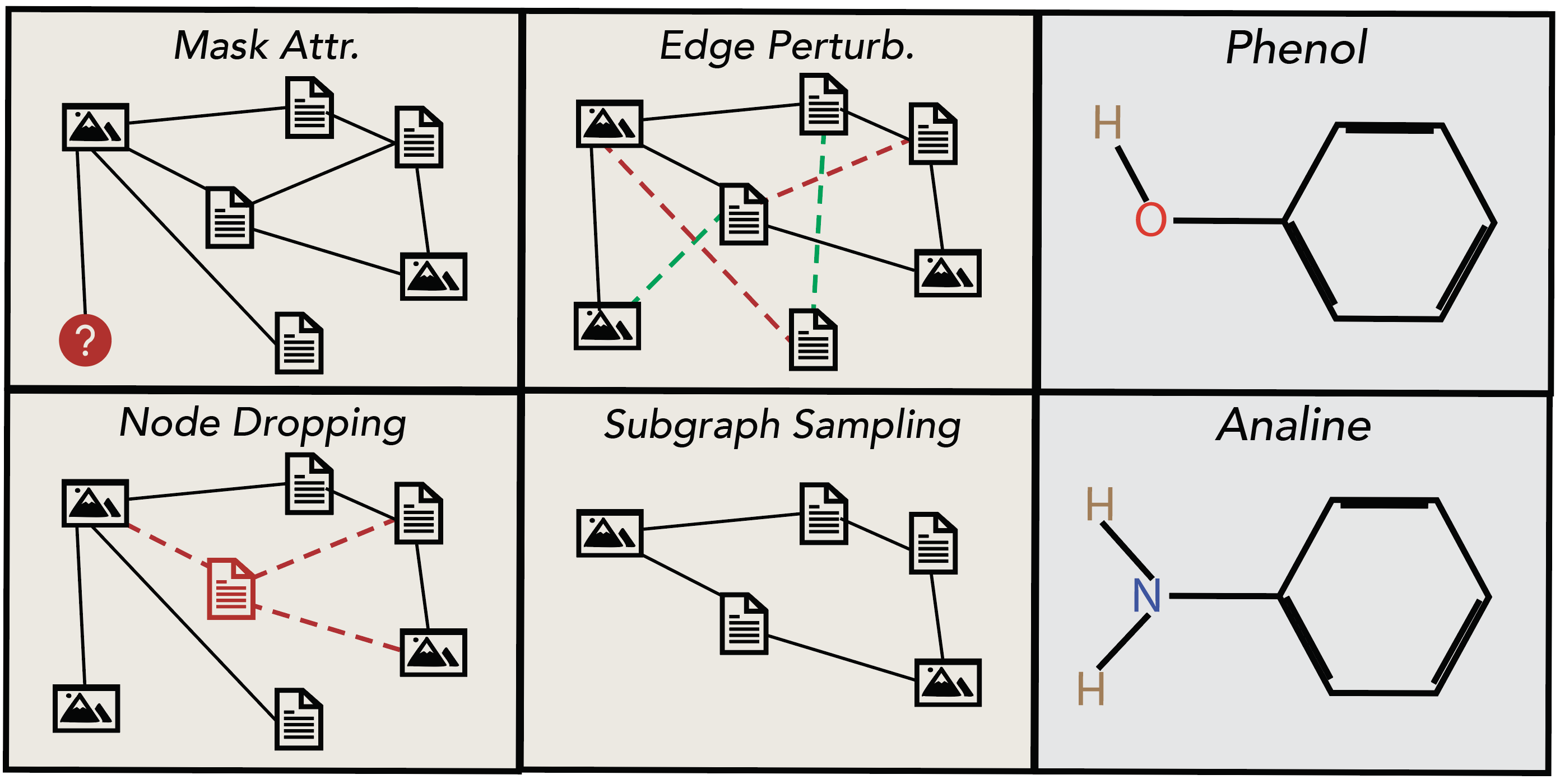}
    \vspace{-0.25cm}
    \caption{[Left] \textit{Domain-Agnostic Graph Augmentations (DAGAs)} introduced in~\cite{You20_GraphContrastiveLearning}. Deletion/addition in red/green. 
    [Right]~\textit{False Positive Samples.} Acidic molecule Phenol and basic molecule Analine are structurally similar but have different properties. DAGAs can inadvertently generate this pair as a positive view, resulting in \textit{similar} representations for semantically \textit{dissimilar} entities.
    }
    \label{fig:molecules}
    \vspace{-0.35cm}
\end{figure}

Interestingly, graph CL (GCL) frameworks often deviate from these key principles and yet report seemingly strong task performance. 
Small, binary graph classification datasets~\cite{Morris20_TU} are routinely used to benchmark GCL frameworks. Moreover, due to the non-euclidean, discrete nature of graphs, it can be difficult to design task-relevant graph data augmentations~\cite{Kong20_FLAG,Zhao20_DataAugGNN} or know what invariances are useful for the downstream task. Therefore, frameworks often rely upon domain-agnostic graph augmentations (DAGAs) ~\cite{You20_GraphContrastiveLearning}. However, DAGAs can destroy task relevant information and yield \emph{invalid/false positive} samples (see Fig.~\ref{fig:molecules}). It is also unclear if DAGAs induce invariances that are useful or semantically meaningful with respect to the downstream task. 

In this work, we investigate the implications of the aforementioned discrepancies by probing the quality of representations learnt by popular GCL frameworks using DAGAs. We show that DAGAs can destroy task-relevant information and lead to weakly discriminative representations. Moreover, on popular, small benchmark datasets, we find that flawed evaluation protocols and the strong inductive bias of GNNs mitigate limitations of DAGAs. 
Our analysis offers several actionable sanity checks and better practices for practitioners when evaluating GCL representation quality. Further, through two case studies on larger, more complex datasets, we demonstrate that task-aware augmentations (TAAs) are necessary for strong performance and discuss how to identify such augmentations amenable to GCL. Our main contributions are summarized as follows:  
\begin{itemize}
    \item\textbf{Analysis of limitations in domain-agnostic augmentations:} We demonstrate that commonly-used DAGAs lead models to learn weakly discriminative representations by inducing invariances to invalid views or false-positives. Across several architectures and datasets, we find these shortcomings are mitigated by the strong inductive bias of GNNs, which allow existing methods to achieve competitive results on benchmark datasets.
    \item\textbf{Identification of methodological flaws \& better practices}: We contextualize recent theoretical work in visual self-supervised learning to identify problematic practices in GCL: (i)~the use of small datasets and (ii)~training with negative-sample frameworks on binary classification datasets. 
    Furthermore, we provide carefully-designed sanity checks for practitioners to assess the benefits of proposed augmentations and frameworks. 
    \item\textbf{Case studies with strong augmentations:} In two case studies on different data modalities, we demonstrate how to leverage simple domain knowledge to develop strong, task-aware graph augmentations. Our systematic process results in up to 20\% accuracy improvements.
\end{itemize}
For reproducibility, our code and data are available at \url{https://github.com/GemsLab/GCLAugmentations-BestPractices}.

%% file: PAGES/020_preliminaries_v3.tex
We begin by introducing CL. We then discuss how strong, task-relevant augmentations and large, diverse datasets underpin the success of VCL. Finally, GCL and graph data augmentation are discussed. Please see \autoref{appendix:related} for additional related work.

\subsection{Contrastive Learning (CL)} 
\mypar{Frameworks \& Losses.} Several CL frameworks~\cite{Chen20_SimCLR,He20_MoCo,HaoChen21_SpecLoss} have been proposed to enforce similarity between positive samples and dissimilarity between negative samples, where positive samples are generated through data augmentation. 
Normalized temperature-scaled cross entropy (NT-XENT) is a popular objective used by several state-of-the-art CL frameworks~\cite{Chen20_SimCLR,Oord18_CPC,Sohn16_MetricLearning,Wu18_ParametricInstanceLevel,You21_GCLA,You20_GraphContrastiveLearning,Suresh21_AdvGCL} and is defined as follows.
Let $\mathcal{X}$ be a data domain, $\mathcal{D}=\{x_{[1\dots n]}|x_i \in \mathcal{X}\}$ be a dataset, $\mathcal{T}\colon \mathcal{X}\rightarrow\tilde{\mathcal{X}}$ be a stochastic data transformation that returns a positive view, and $f\colon\{\mathcal{X},\tilde{\mathcal{X}}\}\rightarrow\mathbb{R}^d$ be an encoder. Further, assume we are given a batch of size $N$, similarity function $\text{sim}\colon(\mathbb{R}^d, \mathbb{R}^d) \rightarrow [ 0,1]$, temperature parameter $\tau$, and encoded positive pair $\{\boldsymbol{z}_i,\boldsymbol{z}_j\}$.
Then, NT-XENT can be defined as: 
\begin{equation}\label{eq:nxtent}
\ell_{i, j}=-\log \frac{\exp \left(\text{sim}\left(\boldsymbol{z}_{i}, \boldsymbol{z}_{j}\right) / \tau\right)}{\sum_{k=1}^{2 N} \mathds{1}_{[k \neq i]} \exp \left(\text{sim}\left(\boldsymbol{z}_{i}, \boldsymbol{z}_{k}\right) / \tau\right)}.\end{equation} 
Here, the numerator encourages the positive pair to be similar, while the denominator encourages negative pairs ($k \neq i$) to be dissimilar. Alternative CL objectives may enforce such (dis)similarity differently (e.g., through margin maximization~\cite{Tschannen20_MIMaxRL} or cosine similarity ~\cite{HaoChen21_SpecLoss}), but the principles discussed below uniformly explain the success of contrastive learning frameworks~\cite{Arora19_TheoreticalAnalysis}.

\vspace{0.15cm}
\mypar{The role of augmentations.}
 Recent work \cite{Tian20_WhatMakesForGoodViews,tsai21_SSLMultiView,Purushwalkam20_DemystifyingCL} has demonstrated that data augmentation is critical for training CL frameworks. Theoretically, Tian et. al ~\cite{Tian20_WhatMakesForGoodViews} show that positive views should preserve task-relevant information, while simultaneously minimizing task-irrelevant information~\cite{tsai21_SSLMultiView}. Training on such views introduces invariances to irrelevant information, leading to more generalizable representations. Indeed, state-of-the-art VCL frameworks~\cite{Chen20_SimCLR, He20_MoCo, DBLP:conf/nips/Caron20_Swav, Chen20_SimSiam, Grill20_BYOL} rely upon strong, task relevant data augmentation to generate such views. For example, Purushwalkam et al.~\cite{Purushwalkam20_DemystifyingCL} show that augmentations used by SimCLR introduce ``occlusion invariance'', which is useful in classification tasks where objects may be occluded. Overall, we highlight that augmentation strategies are not universal~\cite{You20_GraphContrastiveLearning,You21_GCLA} and must align with the task; e.g., semantic segmentation tasks would benefit more from augmentations that induce view-point invariances \cite{Purushwalkam20_DemystifyingCL}.

\vspace{0.15cm}
\mypar{The role of large, high-quality datasets.} Empirically, CL frameworks \cite{Chen20_SimCLR,He20_MoCo,MoCov3} often require many negative samples in each batch to avoid class collisions (i.e., false positives)~\cite{Arora19_TheoreticalAnalysis}. Further, recent theoretical work has shown that optimizing Eq.~(\ref{eq:nxtent}) is equivalent to learning an estimator for the mutual information shared between positive views, where the quality of this estimate is upper-bounded by batch-size \cite{Oord18_CPC, MI_bounds}. These properties combine to necessitate the use of large, diverse datasets in contrastive learning. 

\vspace{-0.15cm}
\subsection{Graph Contrastive Learning (GCL)}
\mypar{Frameworks.} In this paper, we focus on three state-of-the-art unsupervised representation learning frameworks for graph classification that represent different methodological perspectives: 
GraphCL \cite{You20_GraphContrastiveLearning}, InfoGraph~\cite{Sun20_InfoGraph} and MVGRL~\cite{Hassani20_MVGRL}. Similar to SimCLR, GraphCL uses NT-XENT to contrast representations of augmented samples using a shared encoder.
Much like DeepInfoMax~\cite{Hjelm19_DeepInfoMax}, InfoGraph maximizes the mutual information between local and global views, where corresponding views are obtained through subgraph sampling and graph-pooling.  
Meanwhile, MVGRL mirrors CMC~\cite{Tian20_CMC} and uses dual encoders to contrast multiple views of a graph, where views are generated by first running a diffusion process (e.g. Personalized Page Rank ~\cite{Page99_PageRank}, Heat Kernel~\cite{Kondor02_HeatKernel}) over the graph and then sampling subgraphs.

\vspace{0.15cm}
\mypar{Graph data augmentation.} 
Existing GCL frameworks leverage three main strategies to generate views: feature or topological perturbation (GraphCL), sampling (InfoGraph), and/or diffusion processes (MVGRL). 
We focus on the domain-agnostic graph augmentations (DAGAs) introduced by GraphCL, shown in Fig.~\ref{fig:molecules}, as these are more popular in recent frameworks~\cite{You21_GCLA, Thakoor21_GByol, You20_GraphContrastiveLearning}, composable ~\cite{Chen20_SimCLR, Grill20_BYOL}, fast, and do not require dual view encoders. 
An empirical study on the benefits of DAGAs in GCL~\cite{You20_GraphContrastiveLearning} demonstrates that (i) composing augmentations and adjusting augmentation strength to create a more difficult instance discrimination task improves downstream performance and (ii) augmentation utility is dataset dependent.
However, a critical assumption underlying DAGA is that by limiting augmentation strength such that only a fraction of the original graph is modified, task-relevant information is not significantly altered. 
In Sec. \ref{sec:understanding_graph_cl}, we revisit this assumption to show that it does not hold for many datasets and discuss the implications of training with poorly augmented graphs. 
Clearly, it is expected that models trained with task-aware augmentations (TAAs) that induce useful invariances will learn better features than those trained with DAGAs.
However, graphs are often used as abstracted representations of structured data, such as molecules~\cite{Zitnik_Biology} or point clouds~\cite{Shi20_PointGNN}, and it is often unclear how to represent task-relevant invariances after abstracting to the graph space. 
In Sec.~\ref{sec:effective_augmentations}, we discuss a broad strategy for identifying augmentations that induce task-relevant invariances in the abstracted, graph space and demonstrate the significant performance boosts achieved by using such augmentations.

\vspace{0.15cm}
\mypar{Automated Graph Data Augmentation.} Concurrent works \cite{Suresh21_AdvGCL,You21_GCLA,You22_BYOV,Hassani22_LearnAugLearnRep,Kefato21_GraphSurgeon,Kefato21_SelfGNN,Lee21_AugFree,Park21_GraphAugMCMC} have begun investigating \textit{automated} graph data augmentation as a means of both avoiding costly trial and error when selecting augmentations and generating more informative, task relevant views. These methods often use bi-level optimization objectives and/or viewmakers \cite{Tamkin21_Viewmaker} to jointly learn representations and augmentations (cf.\ Appendix~\ref{appendix:related} for more details).  
Our analysis (Sec.~\ref{sec:understanding_graph_cl}) remains pertinent for GCL with automated augmentations.
Namely, the proposed sanity checks are not augmentation specific, the identified evaluation flaws must still be considered, and untrained models should still be included as baselines. 
Also, our discussion on the benefits and properties of TAAs (Sec.~\ref{sec:effective_augmentations}) remains relevant as it is difficult to identify post-hoc if an automated augmentation strategy is inducing semantically meaningful invariances or exploiting shortcuts.

%% file: PAGES/030_revisiting_aug.tex
In this section, we investigate how existing GCL frameworks deviate from the principles underlying the success of VCL methods and the effects of such deviations. 
We discuss and establish three key observations: 
\begin{itemize}
    \item[\bf (O1)] Standard graph data augmentation is susceptible to altering graphs semantics and task-relevant information.
    \item[\bf (O2)] Training on such augmentations can lead to weakly discriminative representations.
    \item[\bf (O3)] The strong inductive bias of randomly-initialized GNNs obfuscates the performance of weak representations and misaligned evaluation practices. 
\end{itemize}

\vspace{0.13cm}
\mypar{Empirical Setup.} In our analysis, 
we focus on commonly used graph classification datasets (Table~\ref{tab:datasets})~\cite{Morris20_TU}.
Official implementations for 
GraphCL\footnote{https://github.com/Shen-Lab/GraphCL}, 
InfoGraph\footnote{ https://github.com/fanyun-sun/InfoGraph}, and 
MVGRL\footnote{https://github.com/kavehhassani/mvgrl} are used.
We consider the encoder architecture used by \cite{You20_GraphContrastiveLearning} 
and report results with graph convolutional layers from 
GIN~\cite{Hu19_HowPowerfulAreGNN} (original implementation), 
PNA~\cite{Corso_PNA}, SAGE~\cite{Hamilton_GraphSage}, GAT~\cite{Velickovic18_GAT}, and GCN~\cite{kipf17_GCN}. 
See \autoref{appendix:2} for details on the training setup.

\begin{table}[h!]
\centering
{
    \vspace{-0.2cm}
    \caption{\textit{Dataset Description}}
    \vspace{-0.2cm}
    \label{tab:datasets}
    \resizebox{\columnwidth}{!}{
    \begin{tabular}{l r@{\hspace{2pt}}  r@{\hspace{2pt}} @{\hspace{2pt}}r @{\hspace{2pt}}r @{\hspace{5pt}}r}
    \toprule
      \textbf{Name} &\textbf{Graphs} & \textbf{Classes} & \textbf{Avg. Nodes} & \textbf{Avg. Edges}  & \textbf{Domain} \\
    \midrule
        IMDB-BINARY \cite{Yanardag15_DeepGraphKernels} & 1000 & 2 & 19.77 & 96.53 & Social\\
        REDDIT-BINARY \cite{Yanardag15_DeepGraphKernels} & 2000 & 2 & 429.63 & 497.75 & Social\\
        GOSSIPCOP \cite{Shu20_FakeNewsNet} & 5464 & 2 & 55.48 & 54.51& News \\
        DEEZER \cite{karateclub} & 9629 & 2 & 23.49 & 65.25 & Social \\
        GITHUB SGZR \cite{karateclub} & 12725 & 2 & 113.79 & 234.64  & Social \\
         MUTAG \cite{Kriege12_SubgraphMatching}  & 188 & 2 & 17.93 & 19.79 & Molecule \\ 
         PROTEINS \cite{Borgwardt05_PROTEINS} & 1113 & 2 & 39.06 & 72.82 & Bioinf. \\
         DD \cite{Shervashidze11_DD}  & 1178 & 2 & 284.32 & 715.66 & Bioinf. \\
         NCI1 \cite{Wale06_NCI1} & 4110 & 2 & 29.87 & 32.30 & Molecule \\ 
    \bottomrule
    \end{tabular}
    }
}
\vspace{-0.3cm}
\end{table}

\begin{figure*}[t]
\centering
        \begin{subfigure}[b]{0.23\textwidth}
                \centering
             \includegraphics[width=\textwidth]{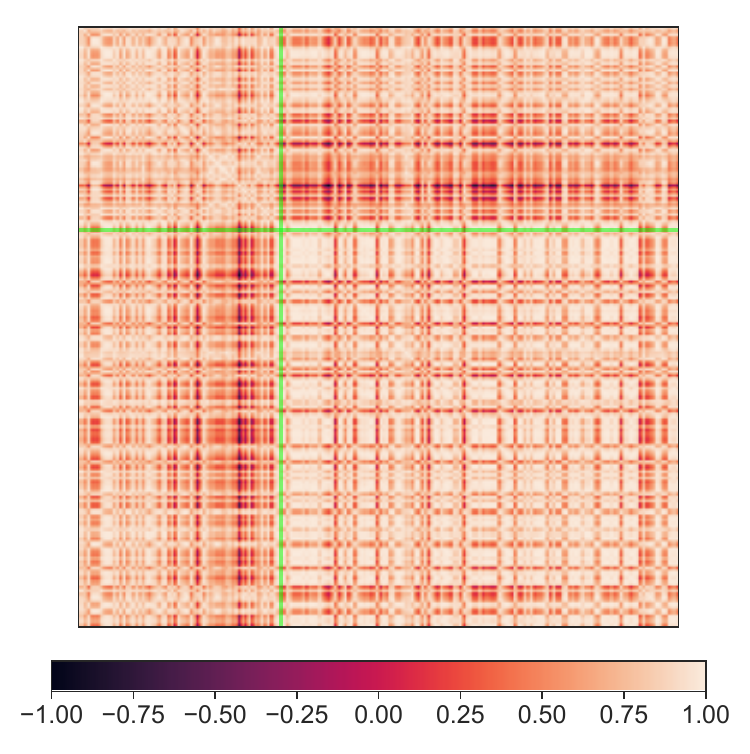}
                \setlength{\abovecaptionskip}{-0.35cm}
                \caption{Random Init. ($85.76 \pm 7.38$)}
                \label{fig:MC_Rand}
        \end{subfigure}%
        \hfill
        \begin{subfigure}[b]{0.23\textwidth}
                \centering
             \includegraphics[width=\textwidth]{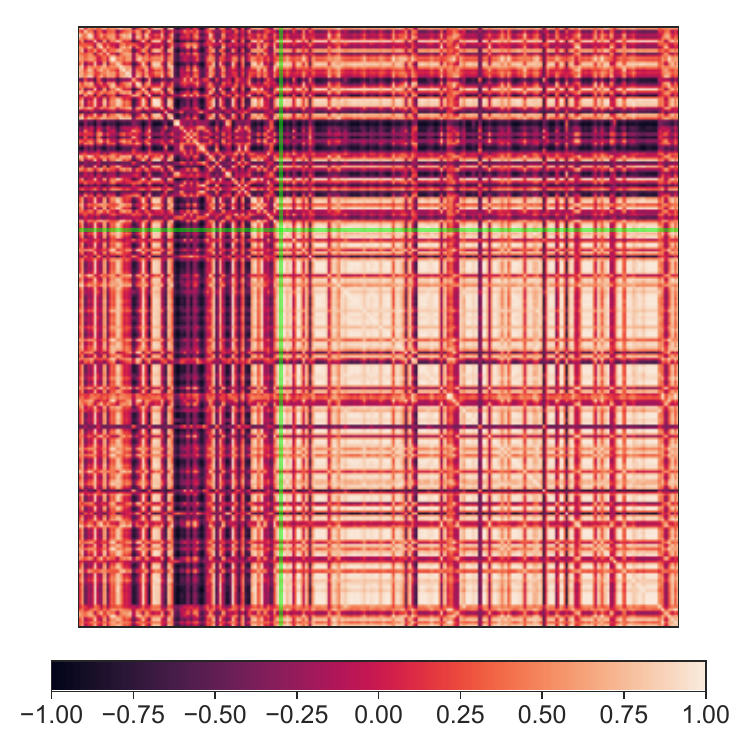}
                \setlength{\abovecaptionskip}{-0.35cm}
                \caption{GraphCL ($86.80 \pm 1.34$)}
                \label{fig:MC_GraphCL}
        \end{subfigure}%
        \hfill
        \begin{subfigure}[b]{0.23\textwidth}
                \centering \includegraphics[width=\textwidth]{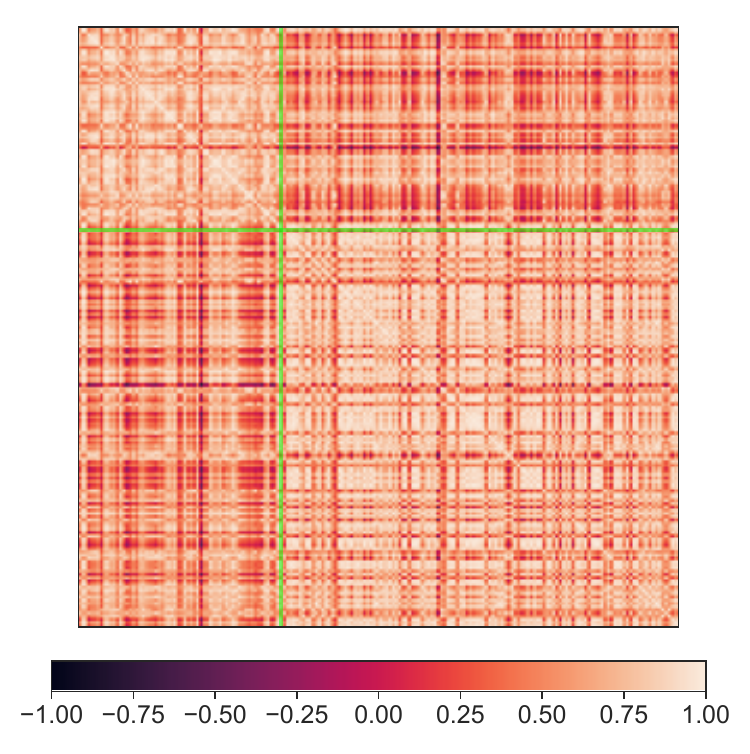}
                \setlength{\abovecaptionskip}{-0.35cm}
                \caption{InfoGraph ($89.01 \pm 1.13$)}
                \label{fig:MC_Infograph}
        \end{subfigure}%
        \hfill
        \begin{subfigure}[b]{0.23\textwidth}
                \centering \includegraphics[width=\textwidth]{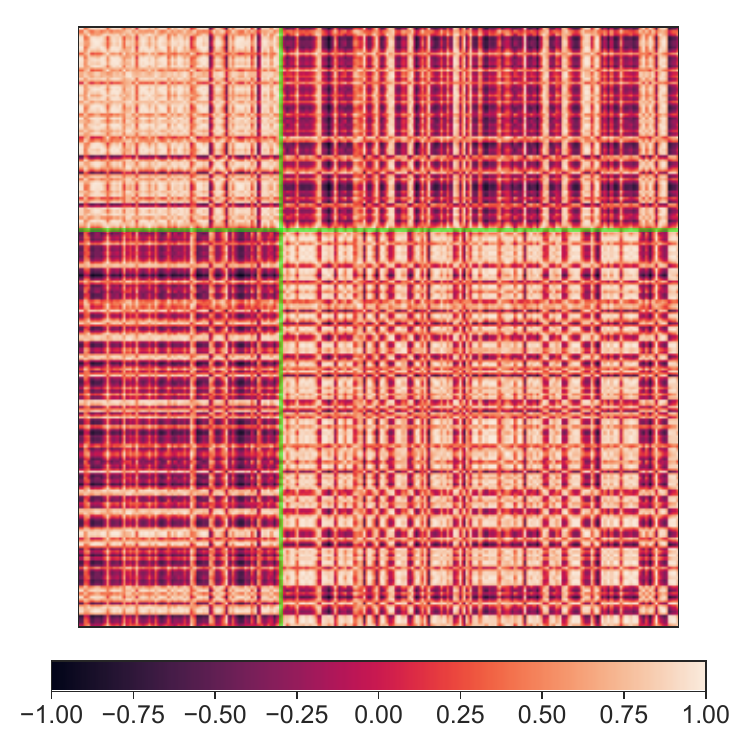}
                \setlength{\abovecaptionskip}{-0.35cm}
                \caption{MVGRL ($89.70 \pm 1.1$)}
                \label{fig:MC_MVGRL}
        \end{subfigure}
        \vspace{-0.3cm}
        \caption{\textit{Representational Similarity.} The normalized cosine similarity between all-pairs of representations is shown above for the MUTAG dataset. The on-diagonal blocks (indicated by green lines) show intra-class similarity, while off-diagonal blocks show inter-class similarity. MVGRL, which uses diffusion-based views, learns representations that have high \textit{intra}-class similarity and low \textit{inter}-class similarity, as desired. InfoGraph, which directly maximizes mutual information between local/global views, preserves high intra-class similarity, and has moderate inter-class similarity. GraphCL, which uses domain-agnostic graph augmentations, has low intra-class similarity in the upper left block. This indicates that training on false positive/invalid samples can negatively impact representational power.} 
        \label{fig:modecollapse}
\end{figure*}

\subsection{(O1) Domain-agnostic graph augmentations alter task-relevant information}\label{sec:l1}
Given the importance of data augmentation in representation learning, several works~\cite{Xiao21_WhatShouldNotBeContrastive,Tian20_WhatMakesForGoodViews,Purushwalkam20_DemystifyingCL, Kugelgen_ProvablySeparates, Zimmermann_Inverts} have investigated its properties. 
Recently, Gontijo-Lopes et al.~\cite{Lopes20_AffinityDiversity} identified an empirical trade-off when selecting amongst augmentations to improve model generalization.
Intuitively, augmentations should generate samples that are close enough to the original data to share task-relevant semantics and different enough to prevent trivially similar samples. This trade-off can be quantified through two metrics, \textit{affinity} and \textit{diversity}. Affinity measures the distribution shift between the augmented and original sample distributions. Diversity quantifies how difficult it is to learn from augmented samples instead of only training samples \cite{Lopes20_AffinityDiversity}. 
While augmentations that best improve generalization optimize for both metrics~\cite{Lopes20_AffinityDiversity}, 
it is not clear that DAGAs also optimize for both. 
For example, molecular graph classification tasks are commonly used to evaluate GCL frameworks. 
However, as noted in Fig.~\ref{fig:molecules}, limited perturbations are needed to invalidate a molecule or significantly alter its function. 
Here, augmented data is sufficiently diverse, but it is not clear if creating invalid molecule samples also leads to low affinity, indicating that task-relevant information have been destroyed. We conduct the following experiment to understand the affinity of DAGAs on benchmark datasets.

\vspace{0.13cm}
\mypar{Experimental setup.} We measure affinity as follows: (i) train a supervised PNA encoder on the original training data, 
(ii) generate an augmented dataset by using random node/subgraph dropping at 20\% of the graph size, as suggested by~\cite{You20_GraphContrastiveLearning} and 
(iii) evaluate on clean \textit{and} augmented training data separately. 
The difference between clean and augmented accuracy quantifies the distribution shift induced by augmentations~\cite{Lopes20_AffinityDiversity}. 

\vspace{0.13cm}
\mypar{Hypothesis.} We argue that while it is not expected that accuracy on augmented data will match that of clean data, augmented accuracy should be nontrivial if augmentations are indeed information-preserving~\cite{Kugelgen_ProvablySeparates, flearning}. 
\vspace{0.13cm}

\begin{table}[t]
    \centering
    \caption{\textit{Augmentation Affinity}. Affinity \cite{Lopes20_AffinityDiversity}, measured by the difference between original and augmented accuracy of a supervised model, captures how much the data distribution has changed as a result of augmentation. We see that DAGAs lead to low affinity. This is expected for molecular datasets, where it is easy to create invalid molecules, and is also true for some social network datasets.}
    \vspace{-0.2cm}
    \label{tab:affinity}
    {\small
    \begin{tabular}{lrr}
    \toprule
    \textbf{Dataset} & \textbf{Clean Train Acc.} & \textbf{Aug. Train Acc.} \\
    \midrule
    MUTAG  & $90.14 \pm 1.36$ & $37.67 \pm 1.48$                                        \\
    PROTEINS         & $70.70 \pm 4.30$ & $56.54 \pm    8.11$                           \\
    NCI1             & $75.55 \pm 4.60$ & $60.15 \pm 0.069$                             \\
    DD               & $84.06 \pm 8.81$ & $65.41  \pm 14.87$                        \\
    REDDIT-BINARY         & $85.56 \pm3.21$ & $50.56 \pm 0.09$                     \\
    IMDB-BINARY          & $70.93 \pm 0.046$ & $50.11 \pm 0.384$  \\
    GOSSIPCOP & $98.047	\pm 0.37$ & $96.03 \pm 1.57$ \\
    \bottomrule
    \end{tabular}
    }
    \vspace{-0.45cm}
\end{table}

\vspace{0.13cm}
\mypar{Results.} In Table \ref{tab:affinity}, we see a considerable difference between clean and augmented accuracy across datasets. This implies low affinity, i.e., a large shift between augmented and training distributions, and confirms that DAGAs can destroy task-relevant information. Consequently, training on such samples will harm downstream task performance, as shown by prior works on VCL~\cite{flearning} and elucidated below for GCL. 
\input{PAGES/031_inductive_bias_table}

\subsection{(O2) Domain-agnostic augmentations induce weak discriminability}\label{sec:l2}

Recall that contrastive losses maximize the similarity between representations of positive pairs while simultaneously minimizing the similarity amongst representations of negative samples. 
However, Obs. (\textbf{O1}) identifies that DAGAs have low affinity, which suggests that task-relevant information has been significantly altered. 
This implies that representation similarity will be maximized for samples that are \textit{not} semantically similar, e.g., false positive samples. 
Consequently, the resulting representations may not be discriminative with respect to downstream classes---i.e., \textit{intra}-class samples may have lower similarity than \textit{inter}-class samples, counter to what is expected. This claim is investigated in the following experiment.

\vspace{0.13cm}
\mypar{Experimental setup.} We measure the discriminative power of representations learned using GCL as follows: given models trained using GraphCL, InfoGraph and MVGRL, we extract representations for the entire dataset. Then, we calculate cosine similarity between all representation pairs. Representational similarity from an untrained model is also included.

\vspace{0.13cm}
\mypar{Hypothesis.} If a model has learned discriminative representations, intra-class similarity should be high while inter-class similarity should be low. 

\vspace{0.13cm}
\mypar{Results.} 
In Fig.~\ref{fig:modecollapse}, we plot the normalized cosine similarity between representations (sorted by class label), such that the upper left and lower right quadrants correspond to the similarity between same-class representations. Results on additional datasets can be found in Appendix~\ref{appendix:3}. We see that MVGRL (Fig.~\ref{fig:MC_MVGRL}) and InfoGraph (Fig.~\ref{fig:MC_Infograph}) are less likely to encounter false positive pairs as they, respectively, use diffusion-based views and maximize mutual information over sampled subgraphs. GraphCL, which uses DAGAs, is more likely to encounter false positive samples that can harm discriminative power (Obs. (\textbf{O1})). 
Correspondingly, MVGRL and InfoGraph both learn representations with higher intra-class similarity than inter-class similarity. 
In contrast, GraphCL has low intra-class similarity as can be seen in the upper-left quadrant (Fig.~\ref{fig:MC_GraphCL}). This implies that the model has not learned features that capture the semantic similarity between the samples belonging to this class. However, we note that while MVGRL has learned discriminative representations, it requires dual encoders and it is unclear what invariances are learnt by training with diffusion-based views. Finally, we find that even though the randomly initialized, untrained model (Fig.~\ref{fig:MC_Rand}) has higher absolute values for average intra- and inter-class similarities than trained methods, it achieves inter-class similarity relatively lower than intra-class similarity, as required for discriminative applications. We further elaborate on this point in the next section. 

\vspace{0.13cm}
\mypar{Proposed evaluation practice.} Given that CL frameworks directly optimize the similarity between representations, we argue that plotting representational similarity can serve as a simple sanity check for practitioners to assess the quality of their model's learned representations. Indeed, models are often only assessed through linear evaluation or task accuracy, which may hide differences in the discriminative power of representations. For example, as shown in Fig.~\ref{fig:modecollapse}, InfoGraph and MVGRL have similar task accuracy, but MVGRL has learnt more discriminative representations. 

\vspace{0.13cm}
Having established that DAGAs can lead to invalid or false positive augmented samples and that training on such samples can lead to poorly-discriminative representations, we next investigate whether other factors are bolstering GCL performance. Specifically, we discuss the role of randomly initialized, untrained GNN inductive bias and identify flaws in current GCL evaluation practices.

\subsection{(O3) Strong inductive bias of random models reduces GCL inefficiencies}\label{sec:l3}
As noted in Obs. (\textbf{O2}), randomly-initialized, untrained GNNs can produce representations that are already discriminative without any training (Fig.~\ref{fig:MC_Rand}). 
While the strength of inductive bias of GNNs in (semi-) supervised settings has been noted before~\cite{kipf17_GCN, zeng09_ComparingStars, Safronov_Untrained-EmbBlog, Suresh21_AdvGCL}, we aim to better contextualize the performance of GCL frameworks by conducting a systematic analysis of the inductive bias of GNNs, using several datasets and architectures. Understanding the performance of untrained models helps contextualize the cost of training. 

\vspace{0.13cm}
\mypar{Empirical setup.} For DEEZER and GITHUB-SGZR, a PNA encoder is used to stabilize training. All other datasets are trained with a GIN encoder. MVGRL ran out-of-memory so we did not include it in this evaluation. See Appendix~\ref{appendix:3} for more details.

\vspace{0.13cm}
\mypar{Results.} As shown in \autoref{tab:statistics}, randomly-initialized, untrained models perform competitively against trained models on several benchmark datasets. It is likely that some of the negative effects of training with DAGAs (Obs. (\textbf{O1})--(\textbf{O2})) were mitigated by this strong inductive bias.
However, note that it becomes difficult to justify the additional cost of GCL on datasets where task performance and representation quality are not noticeably better than untrained models. Below, we discuss how to fairly evaluate GCL frameworks and how popular benchmark datasets are, in fact, inappropriate for GCL.

\vspace{0.13cm}
\mypar{Proposed evaluation practices.} Given that randomly-initialized, untrained models are a non-trivial baseline for GCL frameworks, we argue that they should be included when evaluating novel frameworks to contextualize the benefits of unsupervised training. While some recent works~\cite{Xu21_GraphLog, Suresh21_AdvGCL} include untrained models in their evaluation, this practice remains far from standardized.

Furthermore, CL frameworks often define negative samples through the other samples in the batch. Given the limited size of popular benchmark datasets (Table~\ref{tab:statistics}), it can be difficult to ensure that each batch is large enough to train stably. Further, given that these benchmarks are often binary classification tasks, half the samples, in a balanced setting, are expected to share the positive pair's label but be treated as negative samples. This implies that representations learned with GCL may not be discriminative because models have minimized similarity for semantically related examples. We thus argue that evaluating \textit{GCL} frameworks on these datasets is flawed and this practice should be discontinued. 

We highlight that Dwivedi et al.~\cite{Dwivedi20_BenchmarkingGNNs} also find popular graph classification datasets are problematic in general, but 
for the specific case of GraphCL, this point is of some urgency as such small-scale datasets are part of standard GCL evaluation~\cite{You21_GCLA, Suresh21_AdvGCL}. However, we note that self-supervised frameworks that do not rely on negative samples, such as BYOL\cite{Grill20_BYOL} and SimSiam\cite{Chen20_SimSiam}, can be used as an appropriate alternative for binary datasets. Such frameworks maximize similarity between sample augmentations and avoid degenerate solutions via stop-gradient operations and exponentially moving average target networks.

\subsection{Summary of Proposed Evaluation Practices}
We summarize the practices that we hope will be adopted in future graph CL research:
\begin{itemize}
    \item Given that DAGA can destroy task-relevant information and harm the model's ability to learn discriminative representations, there is need for designing context-aware graph augmentations (Sec.~\ref{sec:effective_augmentations}). 
    \item Randomly initialized, untrained GNNs have strong inductive bias and should be reported during evaluation. 
    \item Small, binary graph datasets are inappropriate for evaluating GCL frameworks. 
    \item GCL frameworks should be comprehensively evaluated using metrics beyond accuracy to assess representation quality.
\end{itemize}

%% file: PAGES/031_inductive_bias_table.tex
\begin{table*}
 \caption{\textit{Inductive Bias on Benchmark Datasets. }Following the same evaluation protocol as \cite{Sun20_InfoGraph}, we generate embeddings from an untrained N-Layer GIN encoder and perform classification using an SVM classifier. Results for GraphCL and InfoGraph are reported from \cite{You20_GraphContrastiveLearning}. Best accuracy is in bold; other models whose accuracy with standard deviation falls within the standard deviation of the best accuracy are underlined. We see across all datasets that untrained models have a strong inductive bias. On PROTEINS, DD, MUTAG DEEZER and GITHUB-SGZR, untrained models perform competitively against trained models.} 
 \label{tab:statistics}
 \centering
 {\small
 \begin{tabular}{l c c  c c  c} 
\toprule
    \cmidrule(r){1-6}
  \textbf{Dataset}  & \textbf{Random Init} & \textbf{Random Init} & \textbf{Random Init}  & \textbf{GraphCL}  & \textbf{InfoGraph}  \\
  (\# Samples)  &  (3 layers) & (4 layers) &  (5 layers) &  \cite{You20_GraphContrastiveLearning} &  \cite{Sun20_InfoGraph} \\
  \midrule
  IMDB-BINARY (1000) & $\underline{67.22 \pm 7.77}$ & $61.26 \pm 7.01$ & $ 60.43 \pm 5.92$  & $71.14 \pm 0.44$ & $\bm{73.03 \pm 0.87}$ \\
  REDDIT-BINARY (2000) & $72.34 \pm 6.64$ & $64.57 \pm 8.03$ &  $67.32 \pm  7.41$ & $\bm{89.53 \pm 0.84}$ & $82.50 \pm 1.42$ \\
 DEEZER (9629) & $ \bm{56.59\pm 0.01}$ & $ \underline{54.99\pm 1.74}$ & $ \underline{54.87 \pm 2.60} $  & $\underline{56.19\pm 0.015}$ &  $ \underline{55.89 \pm 0.88} $ \\
 GITHUB SGZR (12725) & $ 64.51 \pm 0.05$ & $ 64.93\pm 0.04$ & $\underline{64.93 \pm 0.89}$  & $ \bm{65.81 \pm 0.413}$ & Out of Time \\
 \midrule
  MUTAG (188) &$\underline{85.76 \pm 7.38}$ & $\underline{86.36 \pm 6.51}$& $\underline{86.73 \pm 10.33}$  &  $\underline{86.80 \pm 1.34}$  & $\bm{89.01 \pm 1.13}$ \\
  PROTEINS (1113) & $\underline{73.64 \pm 5.464}$ & $\bm{74.46 \pm 4.09}$  & $\underline{74.22 \pm 2.85}$ & $\underline{74.39 \pm 0.45}$ & $\underline{74.44 \pm 0.31}$  \\
  DD (1178) & $\underline{73.23 \pm 8.25}$ & $ \underline{72.15 \pm 7.25}$&  $\underline{77.08 \pm 4.18}$ & $\bm{78.62 \pm 0.40}$ & $72.85 \pm 1.78$ \\
  NCI1 (4110) & $70.65 \pm 1.99$ & $70.36 \pm 3.11 $&  $70.49 \pm 2.42$ &  $\bm{77.81 \pm 0.41}$ & $\underline{76.20 \pm 1.06}$  \\
\bottomrule
\end{tabular}
}
\end{table*}

%% file: PAGES/040_effective_augs_v2.tex
\begin{figure*}[t!]
        \begin{subfigure}[b]{0.33\textwidth}
                \centering
             \includegraphics[width=.90\linewidth]{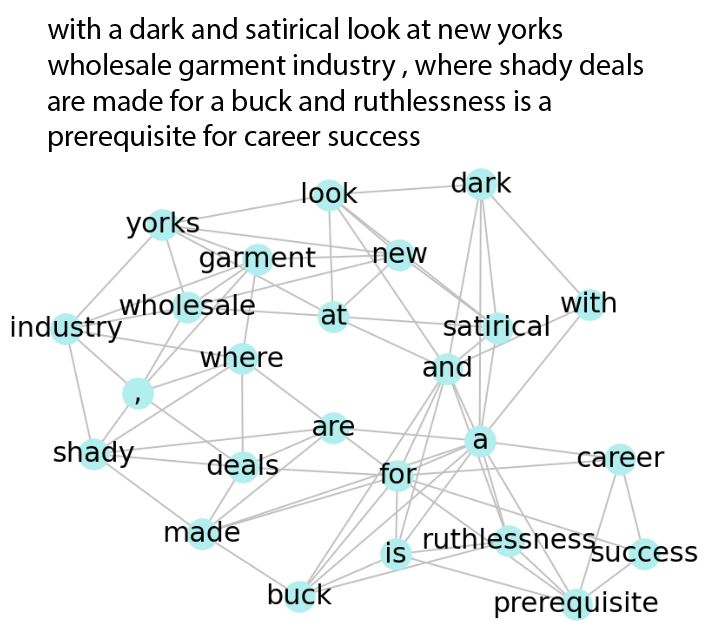}
                \caption{Original}
                \label{fig:words_original}
        \end{subfigure}%
        \begin{subfigure}[b]{0.33\textwidth}
                \centering
             \includegraphics[width=.90\linewidth]{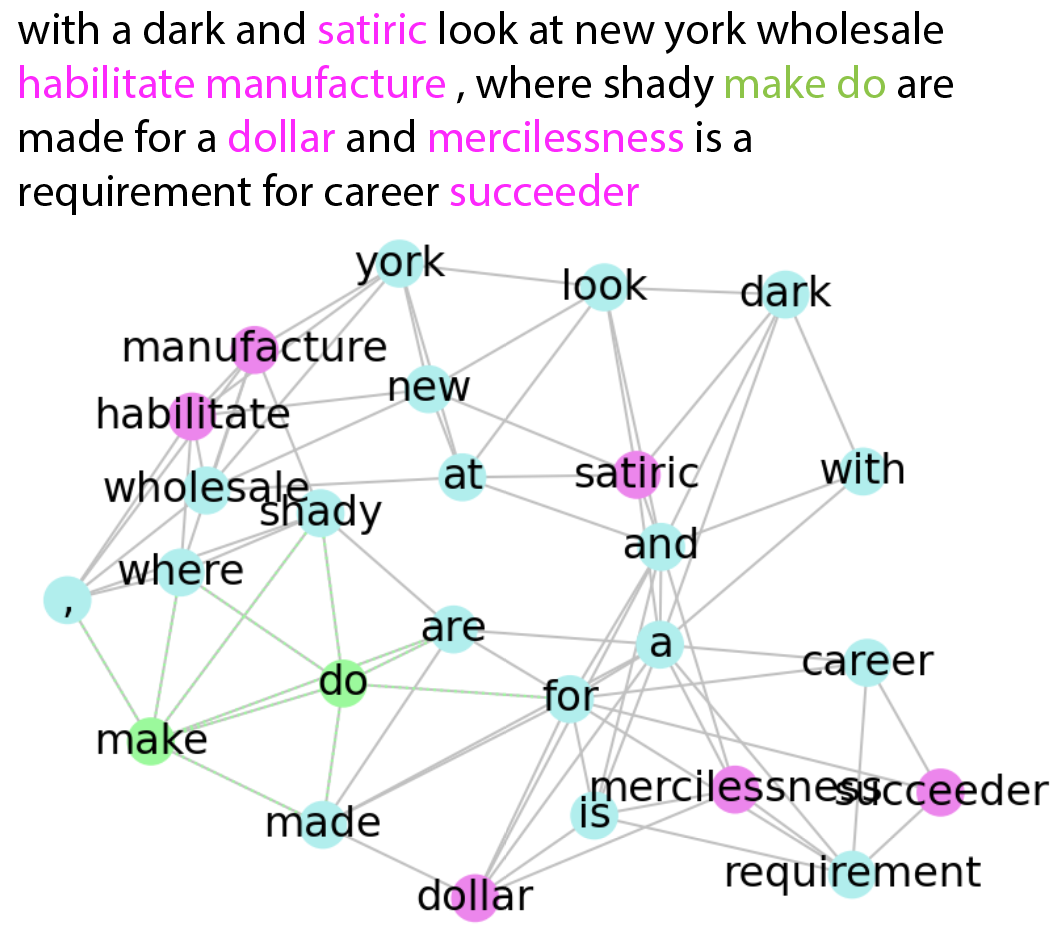}
                \caption{Natural Language Space}
                \label{fig:words_textspace}
        \end{subfigure}%
        \begin{subfigure}[b]{0.33\textwidth}
                \centering \includegraphics[width=.90\linewidth]{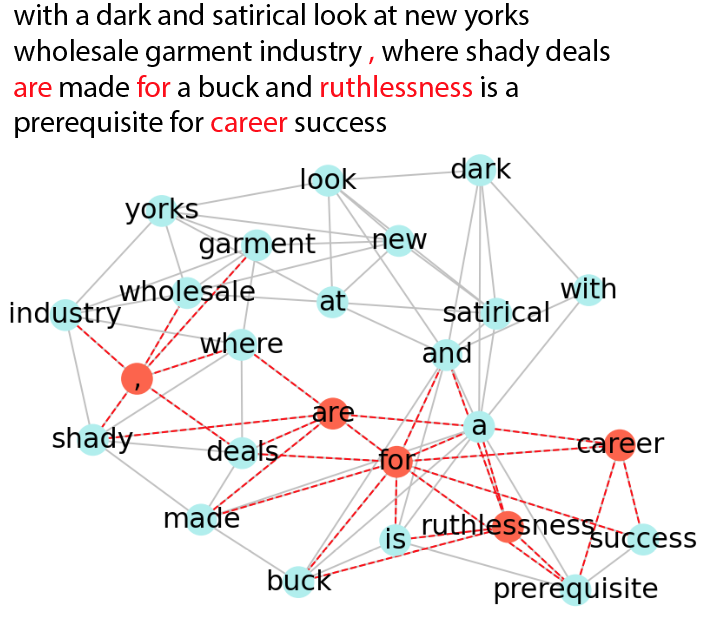}
                \caption{Graph Space}
                \label{fig:words_graphspace}
        \end{subfigure}
         \caption{\textit{Augmentations for Document Classification:} Documents are represented as co-occurrence graphs~\cite{Nikolentzos20_MPAD, Zhang20_TexTING}, where words are treated as nodes with word2vec embeddings, and edges indicate co-occurrence in a sliding windows \cite{Mikolov13_w2vecV1}. As shown in (b), we perform synonym replacement (purple) and random word insertion (green) to augment sentences without losing task-relevant information \cite{Wei19_EDA}. In (c), we show random node (word) deletion (red). Our results show that natural language space augmentations improve classification accuracy substantially over baseline augmentations.}
 \label{fig:words}
\end{figure*} 

\input{PAGES/042_nlp_table} 
In this section, we exemplify the benefits of adhering to key VCL principles by defining a broad strategy for finding task-aware augmentations in scenarios where prior domain knowledge is available. We note the goal of this strategy is not to resolve problematic data augmentations in GCL. Instead, we use the proposed strategy to help elucidate the benefits of abiding by VCL principles in two careful case studies.

\emph{Augmentation strategy:} 
For many graph-based representation learning tasks, structured data, such as documents~\cite{Nikolentzos20_MPAD}, propagation patterns~\cite{monti2019fake}, molecules~\cite{Hu20_PretrainingGNNs}, maps~\cite{Jiang21_TrafficPredictionSurvey}, and point-clouds~\cite{Shi20_PointGNN}, are first abstracted as graphs via a deterministic process before task-specific learning can begin.
In this practical setting, our idea is to leverage knowledge pertaining to the original, structured data to find augmentations that will, in the abstracted graph space, (i) preserve task-relevant information, (ii) break view symmetry, and (iii) introduce semantically meaningful invariance. In our first case study, which focuses on a graph-based document classification task, we achieve this goal by exploiting existing natural language augmentations~\cite{Wei19_EDA} and directly perturbing the raw input before its graph is constructed. However, when given a sufficiently complex graph construction process, it can be unclear if augmentations in the original space will induce useful invariances or retain task-relevant information in the abstracted graph space. In our second case study, which focuses on image classification using super-pixel nearest-neighbor graphs, we encounter this setting and propose to avoid destruction of task-relevant information by deliberately introducing task-\textit{irrelevant} information. We then use augmentations designed to induce invariance to such \textit{irrelevant} information. 
\subsection{Case Study 1: Document Classification}
\label{sec:docuclass}
We first focus on a binary graph-based document classification task. As shown by prior work~\cite{Tao22-webclassification}, graph-based representations are effective at capturing not only the content but also the structure of a document, leading to improved classification performance in this setting.
Here, our goal is to demonstrate adhering to VCL principles by using TAAs is needed to improve task performance.

\vspace{0.13cm}
\mypar{Dataset \& Task.}
The task is to classify movie reviews and plot summaries according to their subjectivity. Following \cite{Nikolentzos20_MPAD}, we convert the Subjectivity document dataset~\cite{Pang04_Subjectivity} (10k samples) into co-occurrence graphs, where nodes represent words, edges indicate that two words have co-occurred within the same window (e.g. window size 2 and 4), and 
node features are word2vec \cite{Mikolov13_w2vecV1} embeddings. 
An example of this conversion is shown in Fig.~\ref{fig:words_original}. Note that we only use positive-view-based self-supervised learning frameworks (e.g., SimSiam, BYOL) because this is a binary classification task (see Sec~\ref{sec:l3}). Accuracy is computed using a $k$NN classifier.

\vspace{0.13cm}
\mypar{Setup of GNN models.} We use a Message Passing Attention Network \cite{Nikolentzos20_MPAD} as the encoder, and a 2-layer MLP as the predictor. The representation dimension is 64, and models are trained using Adam \cite{Kingma14_Adam} with LR=5e-4. Additional training details are given in Appendix~\ref{appendex:docu}. 
We report results with the original GCN layer used by~\cite{Nikolentzos20_MPAD}, as well as with GraphSAGE \cite{Hamilton_GraphSage} and GIN \cite{Hu19_HowPowerfulAreGNN} layers replacing it. 

\vspace{0.13cm}
\mypar{Domain-Agnostic Graph Augmentations.} We conduct an informal grid search to select which DAGAs and augmentation strengths to use. Among node, edge, and subgraph dropping at $\{5\%, 10\%, 20\%\}$ of text length, we find generating both views using subgraph dropping (10\%) performs the best. Generating one view with subgraph dropping (10\%) and the other with node-dropping (10\%) performs second best. We evaluate both strategies.
    
\vspace{0.13cm}
\mypar{Task-Aware Augmentations.}
Recently, Wei et al. \cite{Wei19_EDA} proposed several intuitive augmentations for use in natural language processing, namely: synonym replacement, random word insertion, random word swapping and random word deletion, where the augmentation strength is determined by the sentence length. (See Fig. \ref{fig:words_textspace} for an example.) By design, these augmentations introduce invariances that are useful to downstream tasks (e.g., invariance to the occasional dropped word), preserve task-relevant information, and break view symmetry in the natural language modality. Due to a co-occurrence based construction process, changes in the underlying document will manifest in the corresponding graph, so it is likely that augmentations remain effective for the abstracted space. 

\vspace{0.13cm} 
\mypar{Results.} 
As shown in Table \ref{tab:textclassification}, task-relevant, natural language augmentations perform considerably better (up to +20\%) than domain agnostic graph augmentations for both window sizes. 
Notably, TAAs are necessary to significantly improve performance over an untrained baseline, indicating that adhering to key principles of VCL is indeed beneficial. 

\vspace{0.13cm} 
\mypar{Potential Graph Space Augmentations.}
While natural language augmentations modify samples prior to the graph construction process, it is easy to see that they can be converted into graph augmentations, effectively infusing DAGAs with domain knowledge on how to perturb co-occurence graphs. Specifically, synonym replacement is equivalent to replacing node features of the selected word (node) with the closest word2vec embedding. Random insertion can be approximated in the co-occurence graph by (i) creating a new node with a randomly selected word2vec embedding and (ii) duplicating the connections of an existing node. Random deletion can be represented by (i) randomly removing a node and (ii) rewiring the modified graph to connect neighbors of the removed node. Random swap is equivalent to swapping the features of two nodes. We highlight that domain-agnostic subgraph and node dropping do \textit{not} rewire the co-occurence graph. Thus, it is unclear what invariance to these augmentations represents in the original data modality. In Appendix ~\ref{appendex:docu}, we show that graph-space and document-space synonym replacement perform comparably, but leave the evaluation of other converted graph space augmentations to future work.

\vspace{0.13cm}
In this case study, we were able directly leverage augmentations in the original modality, which are known to preserve task-relevant information and induce useful invariances, to significantly outperform DAGAs. The next case study focuses on a more challenging setting where augmentations in the original modality are not immediately amenable to GCL due to a complex graph construction process and GNN architectural invariances.

\subsection{Case Study 2: Super-pixel Classification}\label{sec:superpix}
Our second case study is based on 
super-pixel MNIST classification, a standard benchmark for evaluating  GNN performance~\cite{Dwivedi20_BenchmarkingGNNs,Knyazev19_UnderstandingAttention}. Here, we pursue an alternative strategy for task-aware augmentation where augmentations must induce invariance to deliberately irrelevant information (e.g., color for digit classification).
\begin{figure}[t]
\begin{minipage}{0.41\columnwidth}
        \begin{subfigure}[b]{\columnwidth}
                \centering
             \includegraphics[width=\columnwidth]{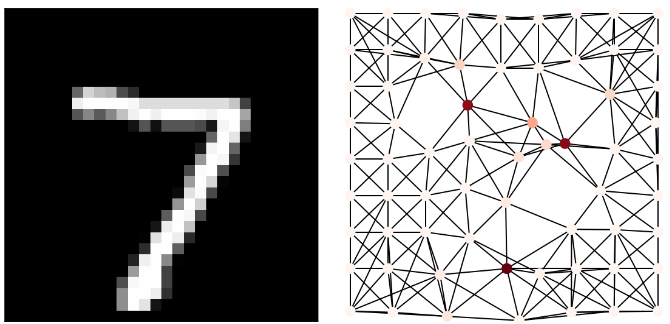}
                \caption{Original}
                \label{fig:mnist_original}
        \end{subfigure}
        \end{minipage}
        \vspace{1em} 
        
        \begin{minipage}{0.41\columnwidth}
        \begin{subfigure}[b]{\columnwidth}
                \centering
             \includegraphics[width=\columnwidth]{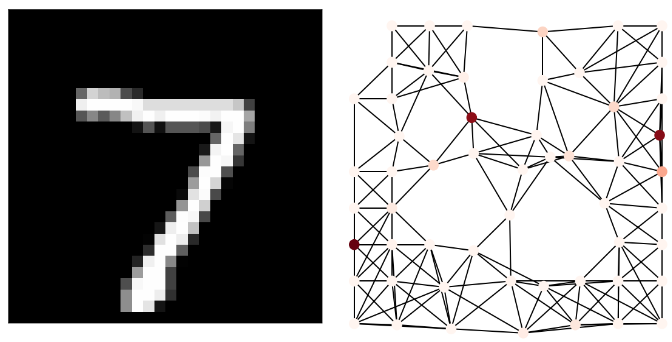}
                \caption{Node Dropping}
                \label{fig:mnist_nodedropping}
        \end{subfigure}
        \end{minipage}
        \hfill
        \begin{minipage}{0.41\columnwidth}
        \begin{subfigure}[b]{\columnwidth}
                \centering
               \includegraphics[width=\columnwidth]{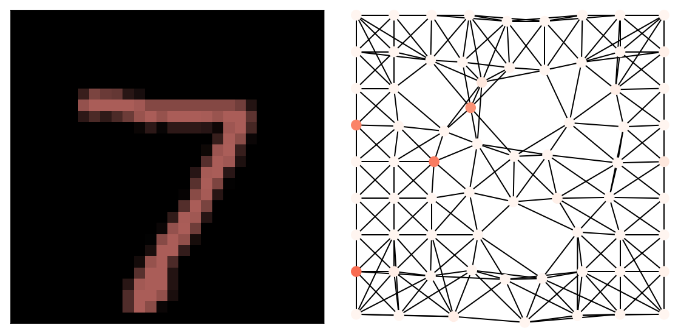}
                \caption{Colorizing}
                \label{fig:mnist_graphspace}
        \end{subfigure}
        \end{minipage}
        \caption{\label{fig:superpixel}\textit{Augmentations for Super-pixel Classification.} Node dropping alters graph topology and it is unclear if task-relevant information is preserved. Colorizing preserves task-relevant information by only perturbing node features.} 
     \label{fig:mnist}
\end{figure}

\vspace{0.13cm}
\mypar{Dataset \& Task.} We follow the established protocols in \cite{Knyazev19_UnderstandingAttention, Dwivedi20_BenchmarkingGNNs} to create super-pixel representations of MNIST, where 
each image is represented as a $k$-nearest neighbors graph between super-pixels (homogeneous patches of intensity). Nodes map to super-pixels, node features are super-pixel intensity and position, and edges are connections to $k$ neighbors. 
An example is shown in Fig.~\ref{fig:superpixel}.

\vspace{0.13cm}
\mypar{Setup of GNN models.} The following architecture is used for experiments. The encoder is 5-layer GIN architecture similar to \cite{You20_GraphContrastiveLearning} and \cite{Dwivedi20_BenchmarkingGNNs}. The projector is a 2-layer MLP and there is no predictor. Models are trained for 80 epochs, using Adam~\cite{Kingma14_Adam} with LR of 1e-3, and the representation dimension is set to 110. 
The models are trained using SimSiam~\cite{Chen20_SimSiam}, BYOL~\cite{Grill20_BYOL}, and SimCLR~\cite{Chen20_SimCLR}. We give more training details in \autoref{appendex:superpix}. 
While composing augmentations is known to improve performance on vision tasks, we avoid it here in order to fairly compare to graph baselines, which only consider a single augmentation. 
\input{PAGES/041_superpixel_table}

\vspace{0.13cm}
\mypar{Domain-Agnostic Graph Augmentations.} Following \cite{You20_GraphContrastiveLearning}, we apply random node dropping at $20\%$ of the graph size to obtain both samples in the positive pair. 

\vspace{0.13cm}
\mypar{Task-Aware Augmentations.} 
While geometric image augmentations \cite{Chen20_SimCLR}, such as horizontal flipping and rotating, generally preserve task-relevant information and introduce semantically meaningful invariance, they cannot break view symmetry in GCL frameworks as GNNs are permutation invariant. Therefore, the representations of a pair of flipped images will be similar as their corresponding super-pixel graph representations are equivalent up to node reordering. On the other hand, augmentations such as cropping may result in qualitatively different super-pixel graphs. Here, it is unclear if the super-pixel graph obtained after augmentation preserves task-relevant information, even if cropping is information preserving with respect to the original image. Therefore, it is not trivial to identify successful augmentations in the abstracted domain that will also be successful in graph space. 

Given the difficulty of identifying augmentations that perturb super-pixel graph topology but also preserve task-relevant information, we focus on image space augmentations that lead to modified node features in the super-pixel graph. Specifically, we select \textit{random colorization} as the TAA as it (i) preserves task-relevant information as color is not relevant property when classifying digits, (ii) breaks view symmetry because the node features of augmented samples are different and (iii) introduces a harmless invariance to colorization. We briefly note that augmentations are generally selected to introduce invariances that are useful to the downstream task. For example, cropping results in occlusion invariance, which is useful for classification tasks where objects are often partially covered \cite{Purushwalkam20_DemystifyingCL}. Here, we take a complementary approach where augmentations introduce harmless information (color) and the model learns to ignore it. This can be a useful strategy when it is difficult to clearly identify potentially useful invariances for a given task.  

\vspace{0.13cm}
\mypar{Results.} 
In Table~\ref{tab:superpixel}, we observe that training with an information-preserving, TAA (colorizing) improves accuracy for both SimSiam and SimCLR. While BYOL generally performs worse than SimSiam and SimCLR, colorizing is still within standard deviation of DAGAs. 
Composing augmentations with colorizing would likely further improve performance, but this investigation is left to future work. This confirms that learning invariance to irrelevant information, as determined by knowledge of the original data modality, is indeed a viable strategy for creating TAAs. Moreover, we note that randomly-initialized models have $37.79\%$ accuracy, indicating that super-pixel data can serve as a sufficiently complex benchmark for future GCL evaluation~\cite{Dwivedi20_BenchmarkingGNNs} (see Appendix~\ref{appendex:superpix} for affinity and representational similarity analysis). 

%% file: PAGES/042_nlp_table.tex
\begin{table*}[t]
\centering
\caption{\textit{Document Classification.} We use domain-agnostic subgraph dropping (S) and node-dropping (N) at 10\% and 5\% of sentence length, respectively, for baseline augmentations. For task-aware augmentations, we stochastically apply synonym replacement (5\%), random insertion (5\%), random swapping (5 \%) and random deletion (10 \%). Random Accuracy with window-size = 2 is $58.46 \pm 1.97$. Random Accuracy with window-size = 4 is $63.93 \pm 0.045$.} 
\vspace{-0.05cm}
\label{tab:textclassification}
{\small
\begin{tabular}{@{}lcrrcrrcrr@{}}
\toprule
&& \multicolumn{2}{c}{\textbf{GCN}} && \multicolumn{2}{c}{\textbf{SAGE}} && \multicolumn{2}{c}{\textbf{GIN}} \\
\cmidrule(lr){3-4} \cmidrule(lr){6-7} \cmidrule(lr){9-10}
\textbf{Augmentation} &&
  \multicolumn{1}{l}{\textbf{SimSiam Acc.}} &
  \multicolumn{1}{l}{\textbf{BYOL Acc.}} &&
  \multicolumn{1}{l}{\textbf{SimSiam Acc.}} &
  \multicolumn{1}{l}{\textbf{BYOL Acc.}} &&
  \multicolumn{1}{l}{\textbf{SimSiam Acc.}} &
  \multicolumn{1}{l}{\textbf{BYOL Acc.}} \\
\midrule
S. vs. S (ws =2)     && $69.41 \pm 7.28$      & $62.98 \pm 3.12$ && $59.17 \pm 8.36 $ & $67.17 \pm 2.70$ && $55.67 \pm 4.61$ & $65.02 \pm 2.00$ \\
S vs. N (ws =2)     && $57.84 \pm  4.31 $ & $65.78 \pm 8.22$ && $56.74 \pm 1.70$       & $63.77 \pm 2.90$      && $58.2 \pm 8.24$  & $74.26 \pm 3.80$    \\
Context-Aware (ws = 2) && $\bm{83.65 \pm 2.31}$  & $\bm{78.12 \pm 2.73}$ && $\bm{81.28 \pm 2.54}$ & $\bm{78.23 \pm 4.53}$ && $\bm{80.37\pm 4.07}$      & $\bm{77.79 \pm 0.09}$      \\
\midrule
S vs. S (ws = 4)    && $61.76 \pm 5.12$ & $66.38 \pm 2.29$      && $54.68 \pm 1.53$ & $67.37 \pm 1.11$      && $54.71 \pm 3.00$ & $66.18 \pm 2.34$    \\
S vs. N (ws = 4)    && $55.38 \pm 1.99$      & $68.31 1.88$      && $59.23 \pm 8.03$  & $70.6 \pm 4.85$      && $53.31 \pm 1.36$     & $66.59 \pm 1.57$      \\
Context-Aware (ws = 4) && $\bm{81.12 \pm 3.97}$    & $\bm{74.05 \pm 5.465}$ && $\bm{80.67 \pm 10.36} $    & $\bm{75.65 \pm 5.54}$      && $\bm{75.30\pm 15.61}$       & $\bm{76.55\pm7.43}$      \\ \bottomrule
\end{tabular}
}
\end{table*}

%% file: PAGES/041_superpixel_table.tex
\begin{table}[t]
    \centering
    \caption{\textit{Super-pixel Classification.} KNN Accuracy after unsupervised training with Node Dropping (ND) or context aware graph augmentations (Colorize) is reported.
    Context aware augmentations improve performance. Accuracy of randomly initialized model is $37.79 \pm 0.03$.}
    \vspace{-0.2cm}
    \label{tab:superpixel}
    {\small
    \begin{tabular}{l r r r} 
    \toprule
      \textbf{Aug.}  & \textbf{SimSiam Acc.} & \textbf{SimCLR Acc.} & \textbf{BYOL Acc.} \\
        \cmidrule(r){1-4}
      ND (20\%) & $66.30 \pm 0.33$ &  $68.56 \pm 0.16 $& $\bm{65.32 \pm 0.95}$\\
      ND (30\%) & $61.30 \pm 0.48$  & $68.07 \pm 0.37$ & $61.87 \pm 1.03$\\
      Colorize & $\bm{68.95 \pm 1.20}$ &  $\bm{73.67 \pm 0.10}$ &  $\underline{64.42 \pm 2.385}$\\
    \bottomrule
    \end{tabular}
    }
\end{table}

%% file: PAGES/050_conclusion.tex
In this work, we discuss limitations in the evaluation and design of existing instance-discrimination GCL frameworks, and introduce new improved practices. In two case studies, we show the benefits of adhering to these practices, particularly the benefits using task-aware augmentations. 
First, through our analysis, we show that domain-agnostic graph augmentations do not preserve task-relevant information and lead to weakly discriminative representations. 
We then demonstrate that benchmark graph classification datasets are not appropriate for evaluating GCL frameworks by contextualizing recent theoretical work in VCL. Indeed, we show that the strong inductive bias of randomly initialized, untrained GNNs obfuscates GCL framework inefficiencies. 
While we acknowledge the community is moving toward larger and more extensive benchmarks~\cite{Dwivedi20_BenchmarkingGNNs}, we emphasize that it is fundamentally incorrect to continue evaluating GCL on legacy graph classification benchmarks. Furthermore, on two case studies with practically complex tasks, we show how to use domain knowledge to perform information-preserving, task-aware augmentation and achieve significant improvements over training with domain-agnostic graph augmentations.
In summary, GCL is an exciting new direction in unsupervised graph representation learning and our work can inform the evaluation of new methods as well as help practitioners design task-aware augmentations. 

%% file: APPENDIX/ObsAll.tex
For Secs. ~\ref{sec:l2},~\ref{sec:l3} experiments, we use a GIN-based encoder \cite{Hu19_HowPowerfulAreGNN} similar to InfoGraph \cite{Sun20_InfoGraph} and GraphCL \cite{You20_GraphContrastiveLearning} for all datasets but (DEEZER, GOSSIPCOP, GITHUB-SGZR). PNA is used for (DEEZER,GITHUB-SGZR) to stabilize Infograph's loss and in Sec.~\ref{sec:l1}. For GOSSIPCOP, the encoder is based off PyG's implementation \cite{Fey19_PyG}: 1 GCN Layer, 1 Linear Layer, embedding dimension = 128, Optimizer = Adam \cite{Kingma14_Adam}, LR = 0.001, \# of Epochs = 25, batch size = 128. 

\vspace{0.05cm}
\noindent \textit{Sec.~\ref{sec:l1} Experimental Setup:}\label{appendix:1} The following training configuration is used: \# of Layers = 3, LR = 0.01, \# of Epochs = 30, Batch-Size = 32. Models are trained on a Nvidia Tesla K80 GPU with Adam. A batch-norm layer is included between the output of the backbone and cross entropy layer. For augmentations, we follow \cite{You20_GraphContrastiveLearning} and stochastically apply node dropping at 20\% of graph size and subgraph dropping at 20\% of graph size.

\begin{figure}[b]
\centering
        \begin{subfigure}[b]{0.3\columnwidth}
                \centering
             \includegraphics[width=\columnwidth]{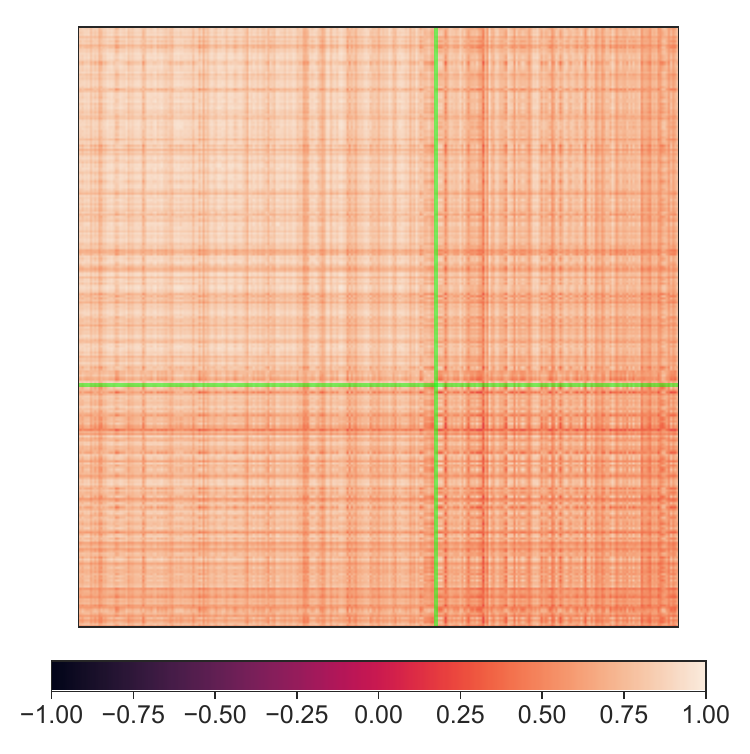}
                \caption{RAND: PROT. \\ $(73.678 \pm 6.91)$}
                \label{fig:app_rand_proteins}
        \end{subfigure}%
        \hfill
        \begin{subfigure}[b]{0.3\columnwidth}
                \centering \includegraphics[width=\columnwidth]{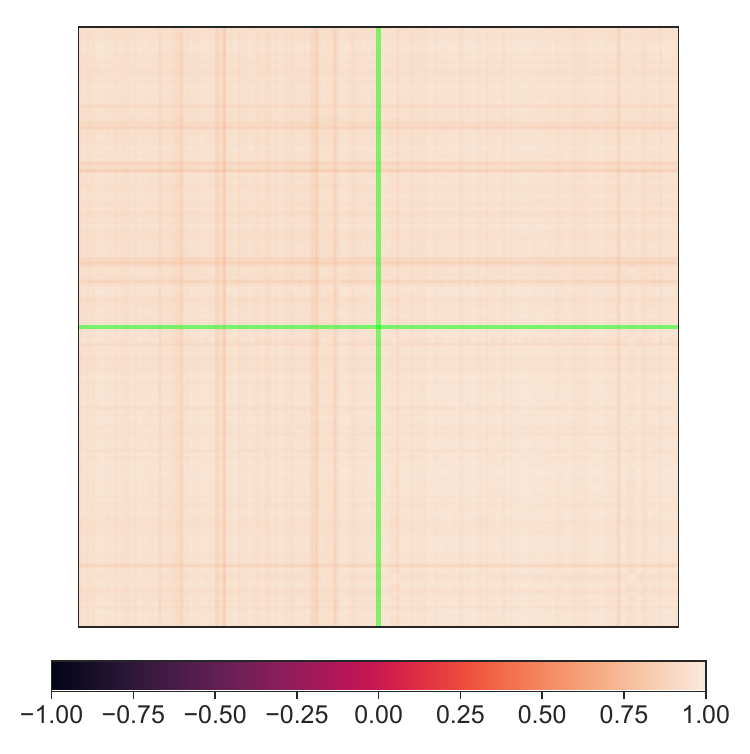}
                \caption{RAND: NCI1 \\ $(70.65 \pm 1.99)$}
                \label{fig:app_rand_nci1}
        \end{subfigure}%
        \hfill
        \begin{subfigure}[b]{0.3\columnwidth}
                \centering \includegraphics[width=\columnwidth]{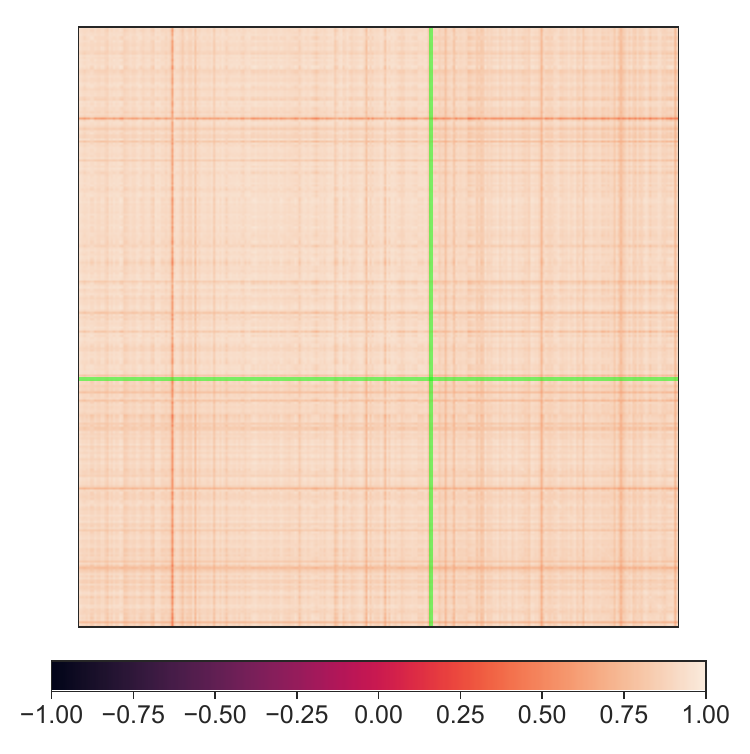}
                \caption{RAND: DD \\ $(74.52 \pm 9.12)$}
                \label{fig:app_rand_dd}
        \end{subfigure}
        \label{fig:modecollapse_rand}
\centering
        \begin{subfigure}[b]{0.3\columnwidth}
                \centering
             \includegraphics[width=\columnwidth]{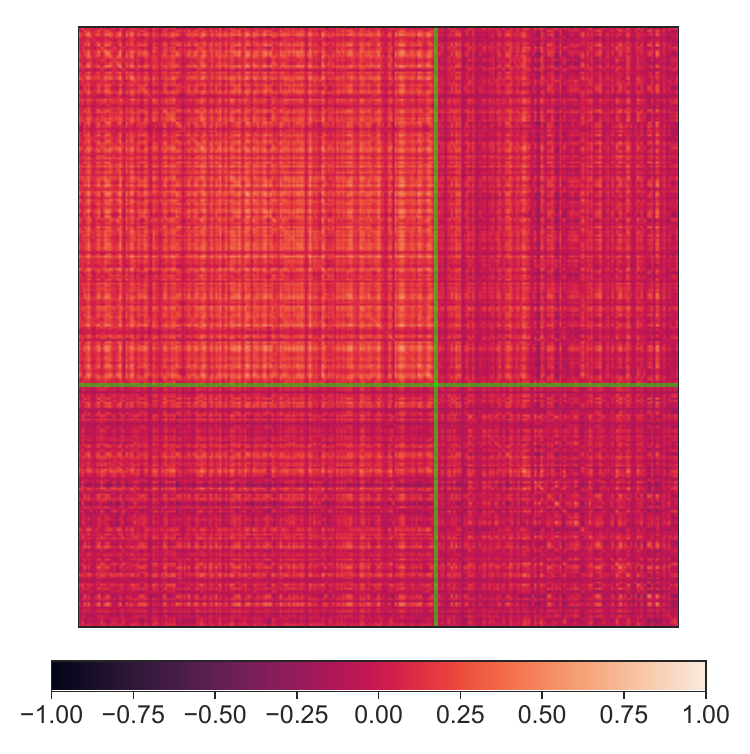}
                \caption{GraphCL: PROT. \\ ($73.49\pm0.33$)}
                \label{fig:app_graphcl_proteins}
        \end{subfigure}%
        \hfill
        \begin{subfigure}[b]{0.3\columnwidth}
                \centering \includegraphics[width=\columnwidth]{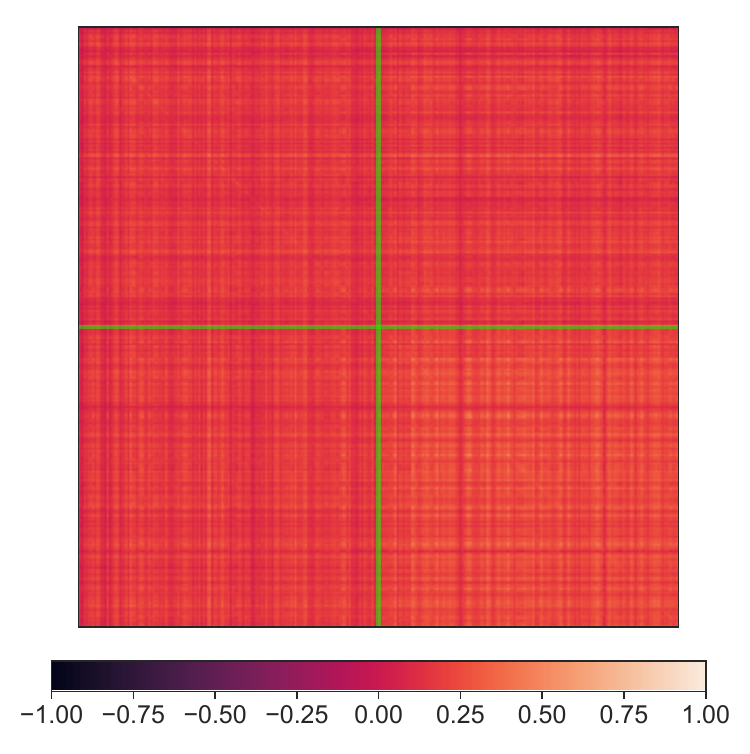}
                \caption{GraphCL: NCI1\\ ($78.16 \pm 0.51$)}
                \label{fig:app_graphcl_nci1}
        \end{subfigure}%
        \hfill
        \begin{subfigure}[b]{0.3\columnwidth}
                \centering \includegraphics[width=\columnwidth]{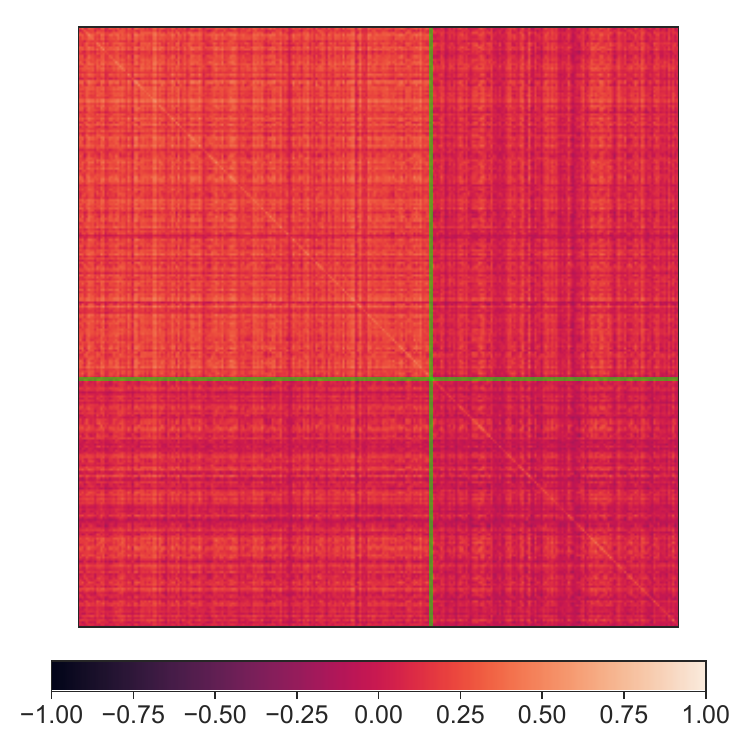}
                \caption{GraphCL: DD \\ $(79.54\pm 0.698)$}
                \label{fig:app_graphcl_dd}
        \end{subfigure}
        \label{fig:modecollapse_graphcl}
\centering
        \begin{subfigure}[b]{0.3\columnwidth}
                \centering
             \includegraphics[width=\columnwidth]{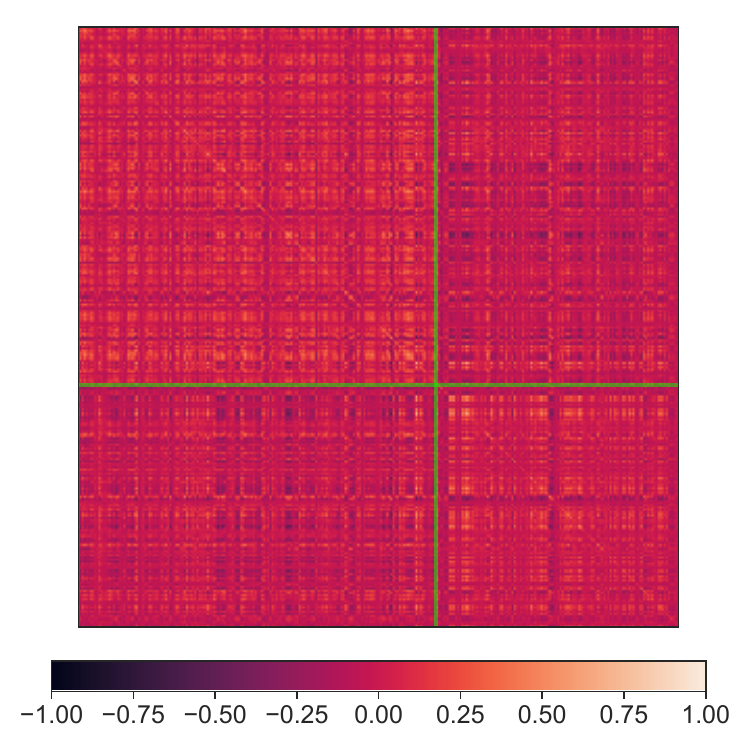}
                \caption{InfoGr: PROT. \\ $(73.225 \pm 0.36)$}
                \label{fig:app_infograph_proteins}
        \end{subfigure}%
        \hfill
        \begin{subfigure}[b]{0.3\columnwidth}
                \centering \includegraphics[width=\columnwidth]{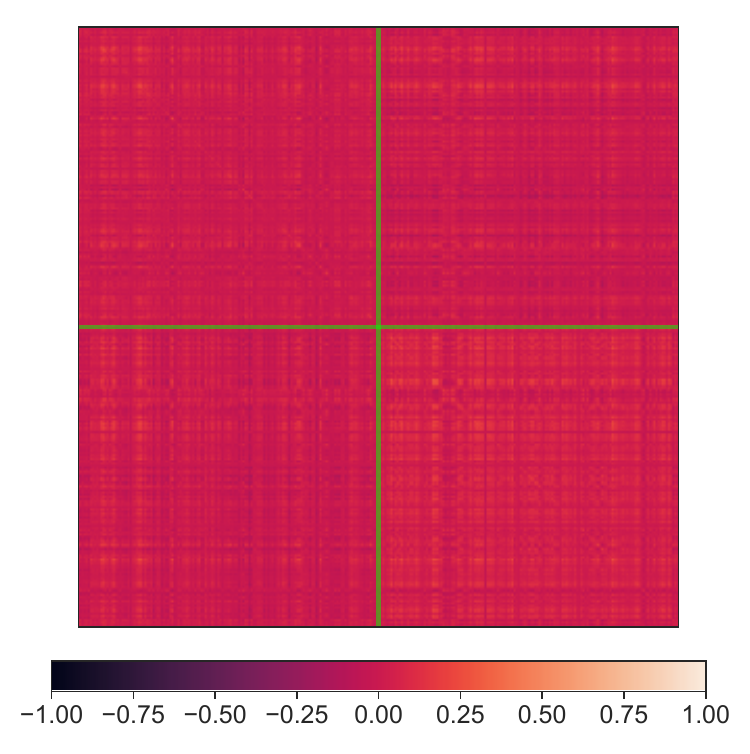}
                \caption{InfoGraph: NCI1 \\ $(73.58 \pm 0.16)$}
                \label{fig:app_infograph_nci1}
        \end{subfigure}%
        \hfill
        \begin{subfigure}[b]{0.3\columnwidth}
                \centering \includegraphics[width=\columnwidth]{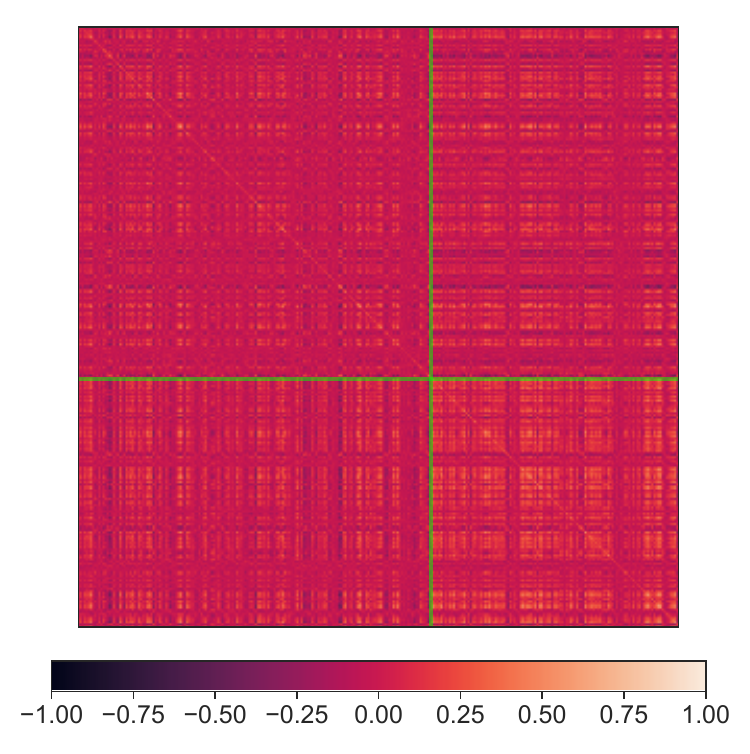}
                \caption{InfoGraph: DD \\ $(69.41 \pm  0.58)$}
                \label{fig:app_infograph_dd}
        \end{subfigure}
        \label{fig:modecollapse_infograph}
    \caption{\textit{Representational Similarity: }In addition to MUTAG (\autoref{fig:modecollapse}), we provide results on PROTEINS, NCI1 and DD. Random inductive bias is most noticeable on MUTAG and PROTEINS. Note that the intra-class similarity can be low for GraphCL and InfoGraph.}
\end{figure}

\vspace{0.05cm}
\noindent \textit{Sec.~\ref{sec:l2} Experimental Setup:}\label{appendix:2} 3-layer GIN model with hidden dimension, learning rate, and epochs trained of (32, NA, NA) for RAND (Random Initialization), (512,0.001,20) for InfoGraph, and (32,0.01,20)  for GraphCL. Adam and Nvidia Tesla K80 GPUs (12-GB GPU) were used to train all models. Results for MVGRL (\cite{Hassani20_MVGRL}) are not included as we consistently witnessed Out-Of-Memory errors. Results are reported over 3 seeds. 
\textit{Additional Results: } Fig.~\ref{fig:app_rand_proteins} includes additional results for PROTEINS, NCI1 and DD datasets. 

\vspace{0.05cm}
\noindent \textit{Sec.~\ref{sec:l3} Experimental Setup: }\label{appendix:3}For all datasets, excluding DEEZER and GITHUB-SGZR, we report results from GraphCL and InfoGraph. We use the same GIN encoder as GraphCL when reporting the performance of randomly initialized models for these datasets. On GITHUB-SGZRS, InfoGraph training time on exceeds eights hours using a NVIDIA Tesla P100. \textit{Additional Results:} See \autoref{tab:append_inductive_bias}. We find that the inductive bias of GNNs is strong across different architectures (GraphSAGE, PNA, GCN, and GAT). 
\input{APPENDIX/random_inductive_bias}

%% file: APPENDIX/random_inductive_bias.tex
\begin{table}[h]
\centering
\caption{\textit{Inductive Bias}: Additional results.}
\label{tab:append_inductive_bias}
\vspace{-0.35cm}
\resizebox{\columnwidth}{!}{%
\begin{tabular}{lllllll}
\toprule
GraphSAGE & 3 Layer & 4 Layer  & 5 Layer  & GraphCL & InfoGraph &  \\
\cmidrule(r){1-7}
MUTAG & $0.85 \pm 0.005$ & $0.85 \pm 0.006$ & $0.85 \pm 0.005$ & $0.82 \pm 0.040$ & $0.85 \pm 0.005$ \\
PROTEINS & $0.73 \pm 0.004$ & $0.73 \pm 0.003$ & $0.74 \pm 0.005$ & $0.75 \pm 0.002$ & $0.74 \pm 0.008$ \\
NCI1 & $0.74 \pm 0.003$ & $0.75 \pm 0.006$ & $0.73 \pm 0.011$ & $0.78 \pm 0.000$ & $0.79 \pm 0.002$ \\
DD & $0.77 \pm 0.006$ & $0.78 \pm 0.002$ & $0.78 \pm 0.005$ & $0.80 \pm 0.008$ & $0.77 \pm 0.010$ \\
REDDIT-B & $0.85 \pm 0.014$ & $0.83 \pm 0.016$ & $0.83 \pm 0.005$ & -- & $0.66 \pm 0.137$ \\
IMDB-B & $0.66 \pm 0.012$ & $0.81 \pm 0.008$ & $0.81 \pm 0.008$ & -- & --  \\
\cmidrule(r){1-7}
PNA & 3 Layer   & 4 Layer   & 5 Layer  & GraphCL  & InfoGraph &  \\
\cmidrule(r){1-7}
MUTAG & $0.88 \pm 0.011$ & $0.88 \pm 0.010$ & $0.89 \pm 0.009$ & $0.86 \pm 0.023$ & $0.90 \pm 0.014$ \\
PROTEINS & $0.74 \pm 0.003$ & $0.74 \pm 0.012$ & $0.74 \pm 0.005$ & $0.74 \pm 0.007$ & $0.74 \pm 0.003$ \\
NCI1 & $0.67 \pm 0.008$ & $0.68 \pm 0.011$ & $0.68 \pm 0.010$ & $0.78 \pm 0.008$ & $0.77 \pm 0.019$ \\
DD & $0.76 \pm 0.014$ & $0.76 \pm 0.002$ & $0.76 \pm 0.008$ & $0.80 \pm 0.008$ & $0.76 \pm 0.006 $ \\
REDDIT-B & $0.90 \pm 0.003$ & $0.88 \pm 0.014$ & $0.89 \pm 0.010$ & $0.92 \pm 0.006$ & $0.92 \pm 0.006$ \\
IMDB-B & $0.72 \pm 0.007$ & $0.68 \pm 0.011$ & $0.68 \pm 0.010$ & $0.71 \pm 0.009$ & $0.71 \pm 0.009$ \\
\cmidrule(r){1-7}
GCN & 3 Layer  & 4 Layer   & 5 Layer   & GraphCL  & InfoGraph &  \\
\cmidrule(r){1-7}
MUTAG & $0.85 \pm 0.003$ & $0.85 \pm 0.004$ & $0.85 \pm 0.005$ & $0.82 \pm 0.013$ & $0.85 \pm 0.003$ \\
PROTEINS & $0.74 \pm 0.003$ & $0.73 \pm 0.007$ & $0.74 \pm 0.004$ & $0.75 \pm 0.004$ & $0.75 \pm 0.003$ \\
NCI1 & $0.76 \pm 0.004$ & $0.75 \pm 0.001$ & $0.75 \pm 0.002$ & $0.78 \pm 0.008$ & $0.79 \pm 0.007$ \\
DD & $0.78 \pm 0.002$ & $0.77 \pm 0.012$ & $0.78 \pm 0.003$ & $0.79 \pm 0.007$ & $0.76 \pm 0.003$ \\
REDDIT-B & $0.52 \pm 0.005$ & $0.51 \pm 0.003$ & $0.52 \pm 0.005$ & $0.92 \pm 0.002$ & $0.80 \pm 0.062$ \\
IMDB-B & $0.54 \pm 0.001$ & $0.57 \pm 0.016$ & $0.58 \pm 0.008$ & $0.71 \pm 0.011$ & $0.62 \pm 0.070$ \\
\cmidrule(r){1-7}
GAT & 3 Layer & 4 Layer &  5 Layer  & GraphCL  & InfoGraph &  \\
\cmidrule(r){1-7}
MUTAG & $0.84 \pm 0.003$ & $0.85 \pm 0.009$ & $0.84 \pm 0.003$ & $0.81 \pm 0.032$ & $0.85 \pm 0.013$ \\
PROTEINS & $0.74 \pm 0.002$ & $0.74 \pm 0.005$ & $0.74 \pm 0.006$ & $0.74 \pm 0.007$ & $0.74 \pm 0.005$ \\
NCI1 & $0.76 \pm 0.009$ & $0.75 \pm 0.004$ & $0.76 \pm 0.002$ & $0.78 \pm 0.004$ & $0.70 \pm 0.040$ \\
DD & $0.78 \pm 0.005$ & $0.77 \pm 0.006$ & $0.79 \pm 0.001$ & $0.79 \pm 0.003$ & $0.76 \pm 0.005$ \\
REDDIT-B & $0.52 \pm 0.005$ & $0.53 \pm 0.004$ & $0.52 \pm 0.012$ & $0.75 \pm 0.004$ & --  \\
IMDB-B & $0.51 \pm 0.004$ & $0.51 \pm 0.009$& $0.50 \pm 0.005 $& $0.51 \pm 0.007$ & -- \\
\bottomrule
\end{tabular}
}
\vspace{-0.35cm}
\end{table}

%% file: APPENDIX/NLP.tex
In Sec.~\ref{sec:docuclass}, we demonstrate the benefits of using task-aware augmentations on a graph-based document classification task. 

\vspace{0.13cm}
\noindent \textit{Experimental Setup:} We use the model, code base and default settings of \cite{Nikolentzos20_MPAD}. Models are trained using Adam: lr = 0.001, weight-decay = 1e-4 and cosine scheduler (T=8). We use the code (https://github.com/ jasonwei20/eda-nlp)  and augmentations by \cite{Wei19_EDA}. Synonym replacement, random deletion, random insertion and random swapping are applied at $5\%, 10\%, 5\%, 5\%$ of sentence length respectively. We generate an augmented version of each sentence for every training epoch. For domain agnostic augmentations, we apply random node dropping ($10\%$) to generate one view. The other view is generated by applying random node 
or subgraph dropping ($10\%$). 

As noted in Sec.~\ref{sec:docuclass}, natural language augmentations can be directly in graph space. We provide proof of concept using the synonym replacement augmentation. In Table~\ref{tab:nlp_appendix}, results are reported for a model trained with synonym replacement and graph space equivalent, node replacement at 5\%. This model achieves comparable accuracy to the original task-aware augmentations.  We suspect that synonym replacement is crucial for this task.

\begin{table}[t]
 \caption{\textit{Document Classification}: We use 
 the same augmentations as in Table~\ref{tab:textclassification}. Text-to-Graph augmentations perform synonym replacement as modifying node features. 
 }
\label{tab:nlp_appendix}
\centering
\vspace{-0.3cm}
\resizebox{.85\columnwidth}{!}{%
 \begin{tabular}{l  c  c } 
\toprule
  \textbf{Augmentation}  & \textbf{(SimSiam) KNN Acc.} & \textbf{(BYOL) KNN Acc.} \\
    \cmidrule(r){1-3}
  S. vs S. (ws = 2)  & $ 62.62 \pm 3.21 $ & $66.25 \pm 2.65$  \\
  S. vs N. (ws = 2) & $ 57.35 \pm 2.47$ & $62.83 \pm 2.82$ \\
  Text-Space (ws = 2) & $ \bm{83.69} \bm{\pm} \bm{0.01}$  & $\bm{82.69 \pm 1.98}$\\
  Text-to-Graph (ws = 2) & $ \bm{83.33} \bm{\pm} \bm{1.29}$  & $78.16\pm2.11$\\

\cmidrule(r){1-3}
  S.  vs S. (ws = 4) & $63.70 \pm 8.71$ & $67.53 \pm 5.00$\\
  S. vs N. (ws = 4) & $54.77 \pm 1.42$  & $65.99 \pm 2.78$\\
  Text-Space (ws = 4) & $\bm{83.29 \pm 0.9}$  & $\bm{72.91 \pm 4.97}$  \\
  Text-to-Graph Space (ws = 4) & $\bm{84.67 \pm 1.57}$  & $\bm{77.96 \pm 2.04}$  \\

\bottomrule
 \end{tabular}
 }
\vspace{-0.2cm}
\end{table} 

%% file: APPENDIX/Superpixel.tex
\begin{table}[b]
\caption{\textit{Comparison to \cite{Verma21_DACL}}. Results only reported for SimCLR, as it performs better than SimSiam and BYOL in preceding experiments.}
\resizebox{0.95\columnwidth}{!}{%
    \begin{tabular}{r r r r r} 
    \toprule
       \textbf{Rand Init.} & \textbf{ND (20\%)} & \textbf{ND (30\%)} & \textbf{Colorize} & \textbf{DACL \cite{Verma21_DACL}}\\
        \cmidrule(r){1-5}
       $37.79 \pm 0.03$ & $68.56 \pm 0.16 $ & $68.07 \pm 0.37$ & $\bm{73.67 \pm 0.10}$ & ${59.94 \pm 0.01}$ \\
    \bottomrule
    \end{tabular}
}
\label{tab:app_dacl}
\end{table}

\begin{table}[b]
\caption{\textit{Super-pixel, Rep. Similarity.} Avg. intraclass and interclass cosine similarity is reported. Colorizing produces representations with the largest difference between intra- vs. inter- class similarity, indicating that representations are well-separated.}
\resizebox{0.95\columnwidth}{!}{
\begin{tabular}{llrrrrr}
\toprule
\textbf{Method} &
  \textbf{Aug.} &
  \textbf{Intra. Sim} &
  \textbf{Inter Sim.} &
  \textbf{Abs. Diff} &
  \textbf{Rel. Diff} &
  \textbf{Acc.} \\
\cmidrule(r){1-7}
SimCLR & ND (20\%) & 86.671 & 78.622 & 8.04   & 0.0928 &  $68.56 \pm 0.16 $ \\
SimCLR & ND (30\%) & 87.03  & 79.05  & 7.987  & 0.091 & $68.07 \pm 0.37$  \\
SimCLR & Colorizing  & 80.801 & 67.812 & 12.988 & 0.1607 & $\bm{73.67 \pm 0.10}$  \\
\bottomrule
\end{tabular}
}
\label{tab:app_superpix_sim}
\end{table}

\begin{table}[b]
\small
 \caption{\textit{Super-pixel Affinity.} Supervised, clean train accuracy is 90.01\% and clean test accuracy is 88.69\%.}
\begin{tabular}{lrr}
\toprule
\textbf{Aug.} & \textbf{Aug. Train Acc. }    & \textbf{Aug. Test Acc.}    \\
\cmidrule(r){1-3}
ND (20\%)    & $39.42 \pm 0.011$ & $40.29 \pm 0.054$ \\
ND (30\%)    & $29.19 \pm 0.01$ & $29.09 \pm 0.036$ \\ 
Colorizing   & \bm{$47.86 \pm 0.05$} & \bm{$48.97 \pm 0.03$} \\
\bottomrule
\end{tabular}
\label{tab:app_affinity_superpix}
\end{table}

In  Sec.~\ref{sec:superpix}, we demonstrate the benefits of using task-aware augmentations  via a case study on MNIST superpixel classification. 

\vspace{0.05cm}
\noindent \textit{Experimental Setup: }50K images are used for training, 10K for validation, and 10K for testing. We follow the same procedure as \cite{Dwivedi20_BenchmarkingGNNs} to convert images to superpixel graphs: SLIC (\cite{Achanta12_SLIC}) is used to extract superpixels from the image. Then, a $k$NN graph is constructed between the superpixels. Node features are RGB values and $(x,y)$ coordinates of superpixels.  Classification is performed using three CL frameworks: SimSiam (\cite{Chen20_SimSiam}), SimCLR (\cite{Chen20_SimCLR}), and BYOL (\cite{Grill20_BYOL}). The same hyper-parameters and architecture are used for all frameworks. Specifically, we use a 5-Layer GIN model closely following \cite{Dwivedi20_BenchmarkingGNNs}. This model is converted from DGL (https://www.dgl.ai) to PyG (\cite{Fey19_PyG}). The following hyper-parameters are used: LR=5e-4, Hidden-Dim =110, Epochs=80, Batch-size = 128. The Adam (\cite{Kingma14_Adam}) Optimizer is used for training. The projector is a 2-layer MLP. The predictor is a 2-layer MLP. Predictor hidden dimension is 1028. Bottleneck dimension is 128. Results are reported over 3 seeds. DAGAs are random node dropping (at 20\% and 30\%). The task-aware augmentation is random colorizing, performed using Scikit-Image (\cite{scikit-image}). As discussed in the main text, colorizing can be represented as transformation on node features as well.

\vspace{0.05cm}
\noindent \textit{Additional Results:} \cite{Verma21_DACL} proposes to mix-up samples at either the input or hidden representation level as an alternative to domain-specific augmentations. However, we find that \cite{Verma21_DACL} under-performs both node-dropping and colorizing, despite tuning the mixing parameter, $\alpha$ (see Table. ~\ref{tab:app_dacl}).This indicates that context-aware and topological augmentations are still important to GCL. Table~\ref{tab:app_superpix_sim} shows intra/inter similarity and Table~\ref{tab:app_affinity_superpix} shows the affinity.

%% file: APPENDIX/related_work.tex
\begin{table}[b]
  \caption{\textit{Selected GCL Frameworks}}
  \label{tab:relatedworks}
  \label{sample-table}
  \centering
  \vspace{-0.3cm}
  \resizebox{\columnwidth}{!}{%
  \begin{tabular}{l p{10cm}}
    \toprule
    \cmidrule(r){1-2}
    Method & Augmentations\\
    \midrule
    BGRL \cite{Thakoor21_BGRL} & Edge Dropping, Attr. Masking \\
    GCA \cite{Zhu20_GCA} & Edge Dropping, Attr. Masking (both weighted by centrality) \\
    GCC \cite{Qiu20_GCC} &  RWR Subgraph Extraction of Ego Network \\
    GraphCL \cite{You20_GraphContrastiveLearning} & Node Dropping, Edge Adding/Dropping, Attr. Masking, Subgraph Extraction    \\
    MVGRL \cite{Hassani20_MVGRL} & PPR Diffusion + Sampling\\
    SelfGNN \cite{Kefato21_SelfGNN}  & Attr. Splitting, Attr. Standardization + Scaling, Local Degree Profile, Paste + Local Degree Profile \\
    JOAO \cite{You21_GCLA} & Min-Max Optimization to adaptively and dynamically select from DAGA set \\
    GraphSurgeon \cite{Kefato21_GraphSurgeon} & Learnable Feature Augmentors that can be applied pre/post encoding\\
    BYOV \cite{You22_BYOV} & Uses graph generation (regularized by InfoMin + InfoBottleNeck) as viewmaker \\
    AdvGCL \cite{Suresh21_AdvGCL} & Adversarial/MinMax Optimization over learnable augmentations\\
    AF-GRL \cite{Lee21_AugFree} & Finds node-level positive samples sharing ``local structure and global semantics''\\
    LG2AR \cite{Hassani22_LearnAugLearnRep} & Learns a policy over augmentations and their respective strengths without bi-level optimization \\
    \bottomrule
  \end{tabular}
  }
\end{table}

\textit{Graph Data Augmentation. } 
\cite{Zhao20_DataAugGNN} train a neural edge predictor to increase homophily by adding edges between nodes expected to be of the same class and break edges between nodes of expected dissimilar classes. However, this approach is expensive and not applicable to graph classification. \cite{Kong20_FLAG} focus on feature augmentations because it is easier than designing information preserving topological transformations. They add adversarial perturbations to node features as augmentations. In unsupervised settings, labels are not available and cannot be used for the adversarial perturbation, so the proposed approach is not directly applicable.
Since the writing of this paper, several recent works have been proposed that perform automatic data-augmentation, some of which we briefly describe in Table \ref{tab:relatedworks}. 

\textit{Graph Self-Supervised Learning. }Several paradigms for self-super-vised learning in graphs have been recently explored, including the use of pre-text tasks, multi-tasks, and unsupervised learning. See \cite{Liu21_SSLGNNSurvey} for an up-to-date survey. Graph pre-text tasks are often reminiscent of image in-painting tasks \cite{Yu18_InPainting}, and seek to complete masked graphs and/or node features (\cite{You20_WhenDoesSSLHelp,Hu20_PretrainingGNNs}). Other successful approaches include predicting graph level or property level properties during pre-training or part of regular training to prevent overfitting (\cite{Hu20_PretrainingGNNs}). These tasks often must be carefully selected to avoid negative transfer between tasks. Many unsupervised approaches have also been proposed. \cite{Sun20_InfoGraph, Velickovic19_DGI} draw inspiration from \cite{Hjelm19_DeepInfoMax} and maximize the mutual information between global and local representations; MVGRL (\cite{Hassani20_MVGRL}) contrasts different views at multiple granularities similar to \cite{Oord18_CPC}; \cite{You20_GraphContrastiveLearning, Qiu20_GCC, Zhu20_GCA,Thakoor21_BGRL, Kefato21_SelfGNN} use augmentations to generate views for contrastive learning. See Table \ref{tab:relatedworks} for a summary of the augmentations used. 